\documentclass[manuscript,screen]{acmart}
\AtBeginDocument{%
  }

\setcopyright{acmlicensed}
\copyrightyear{2018}
\acmYear{2018}
\acmDOI{XXXXXXX.XXXXXXX}

\acmConference[Conference acronym 'XX]{Make sure to enter the correct
  conference title from your rights confirmation emai}{June 03--05,
  2018}{Woodstock, NY}
\acmISBN{978-1-4503-XXXX-X/18/06}

\usepackage{algorithm2e}
\usepackage{float}

\begin{document}

\title{AMSnet-KG: A Netlist Dataset for LLM-based AMS Circuit Auto-Design \\Using Knowledge Graph RAG}

\author{Yichen Shi}
\affiliation{%
  \institution{Eastern Institute of Technology}
  \city{Ningbo}
  \country{China}
}
\authornote{Both authors contributed equally to this research. Address comments to Lei He at Lei.Hexun@gmail.com.}

\author{Zhuofu Tao}
\authornotemark[1]
\affiliation{%
  \institution{University of California, Los Angeles}
  \city{Los Angeles}
  \country{USA}
}
\author{Yuhao Gao}
\affiliation{%
  \institution{BTD Inc.}
 \city{Ningbo}
 \country{China}
  }

\author{Tianjia Zhou}
\author{Cheng Chang}
\author{Yaxing Wang}
\author{Bingyu Chen}
\author{Genhao Zhang}
\affiliation{%
  \institution{University of California, Los Angeles}
  \city{Los Angeles}
  \country{USA}
}


\author{Alvin Liu}
\affiliation{%
  \institution{BTD Inc.}
 \city{Shenzhen}
 \country{China}
}


\author{Zhiping Yu}
\affiliation{%
  \institution{Tsinghua University}
  \city{Beijing}
  \country{China}
  }

\author{Ting-Jung Lin}
\affiliation{%
  \institution{
  Institute of Digital Twin, Eastern Institute of Technology}
  \city{Ningbo}
  \country{China}
  }

\author{Lei He}
\affiliation{%
  \institution{University of California, Los Angeles}
  \city{Los Angeles}
  \country{USA}
  }
\email{lhe@ee.ucla.edu}
\begin{abstract}


High-performance analog and mixed-signal (AMS) circuits are mainly full-custom designed, which is time-consuming and labor-intensive. A significant portion of the effort is experience-driven, which makes the automation of AMS circuit design a formidable challenge. Large language models (LLMs) have emerged as powerful tools for Electronic Design Automation (EDA) applications, fostering advancements in the automatic design process for large-scale AMS circuits. However, the absence of high-quality datasets has led to issues such as model hallucination, which undermines the robustness of automatically generated circuit designs. To address this issue, this paper introduces AMSnet-KG, a dataset encompassing various AMS circuit schematics and netlists. We construct a knowledge graph with annotations on detailed functional and performance characteristics. Facilitated by AMSnet-KG, we propose an automated AMS circuit generation framework that utilizes the comprehensive knowledge embedded in LLMs. We first formulate a design strategy (e.g., circuit architecture using a number of circuit components) based on required specifications. Next, matched circuit components are retrieved and assembled into a complete topology, and transistor sizing is obtained through Bayesian optimization. Simulation results of the netlist are fed back to the LLM for further topology refinement, ensuring the circuit design specifications are met. We perform case studies of operational amplifier and comparator design to verify the automatic design flow from specifications to netlists with minimal human effort. The dataset used in this paper will be open-sourced upon publishing of this paper.


\end{abstract}

\ccsdesc[500]{Hardware~Design databases for EDA}


\keywords{EDA, LLM, AMSnet, Knowledge Graph, RAG, Topology Design}

\maketitle

\section{Introduction}
Digital circuit synthesis has been extensively utilized in electronic design automation (EDA) \cite{michel1992synthesis}, enabling contemporary large-scale digital integrated circuits (ICs) in accordance with Moore's Law. However, the degree of automation in analog and mixed-signal (AMS) circuit design significantly lags behind that of digital design \cite{rutenbar2015analog}. Today's AMS circuits are primarily full-custom designed and still heavily depend on human experts to determine circuit topologies and component sizing. The time-consuming and labor-intensive design process significantly hinders the scalability of AMS circuits.

Large language models (LLMs) have recently demonstrated the potential to address various EDA challenges\cite{zhong2023llm4eda}, offering new hopes for automatic AMS circuit design. However, the availability of high-quality digital circuit data on the Internet, such as Verilog code, far exceeds that of AMS circuits mainly represented by SPICE netlists \cite{lai2024analogcoder}. Consequently, LLMs have been extensively applied in the digital circuit domain, with significant work done in RTL code generation\cite{lu2024rtllm, thakur2023rtlbenchmarking}, yielding promising results. However, their application in AMS circuits is still in an exploratory phase. Researchers have utilized LLMs to assist analog circuit topology design and transistor sizing\cite{lai2024analogcoder, Artisan, yin2024ado, liu2024ladac}. Compared to traditional AI methods such as reinforcement learning and Bayesian optimization (BO), LLMs demonstrate much better interactivity, knowledge transfer and expansion capabilities, and interpretability of design solutions. 


However, due to the scarcity of public AMS circuit netlist datasets, LLMs are not sufficiently trained to produce accurate netlists with correct topology and reasonable sizing\cite{Artisan, lai2024analogcoder}. Researchers have improved the performance of LLMs in designing AMS circuits by training them with datasets collected for specific circuits \cite{Artisan}, or by employing prompt engineering techniques including chain of thought (CoT) \cite{cot}, ReAct \cite{yao2022react}, and in-context learning (ICL) \cite{icl1,icl2,lai2024analogcoder}. Similarly, retrieval-augmented generation (RAG) is also used to supplement LLMs with additional knowledge without requiring the expensive retraining \cite{liu2024ladac}.


\begin{figure}[!ht]
    \centering
    \includegraphics[width=\textwidth]{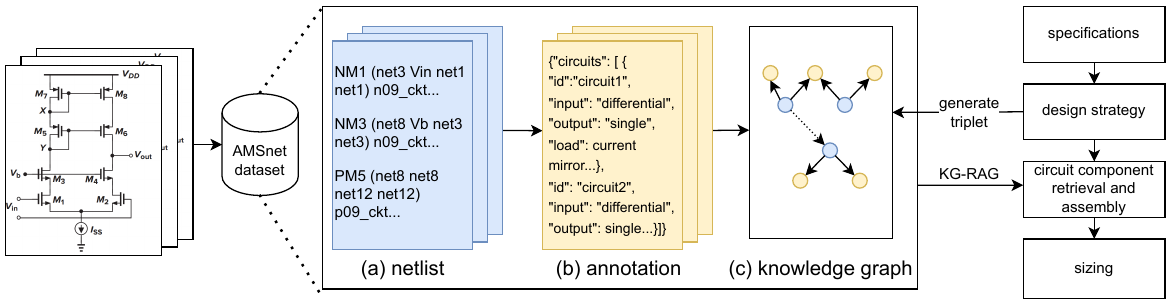}
    \caption{AMSnet-KG dataset (left \& middle) and the workflow diagram of automatic AMS circuit topology design (right). AMSnet-KG dataset includes (a) circuit netlists, (b) corresponding annotations, and (c) an overall knowledge graph encompassing all circuits. Given the specifications, leveraging the extensive knowledge of LLM, the flow extracts relevant relation query triplets from the responses and retrieves the corresponding circuit topology from the knowledge graph. Automatic testbench pairing and transistor sizing complete the circuit design flow.}
    \label{fig:overall}
\end{figure}

\begin{figure}[!ht]
    \centering
    \includegraphics[width=\textwidth]{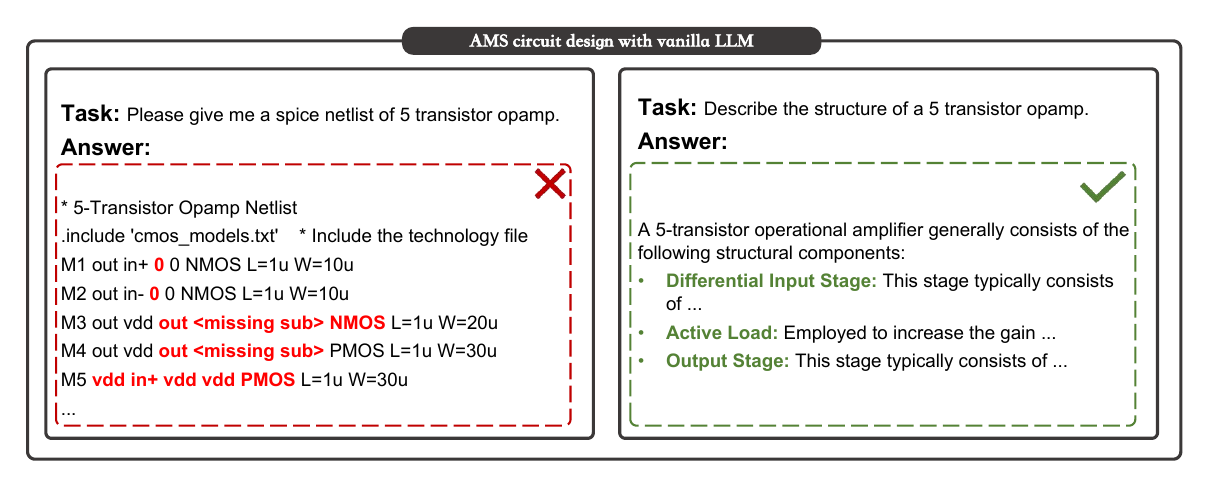}
    \caption{LLM generating the full netlist directly versus describing its building blocks}
    \label{fig:wrong_right}
\end{figure}

Although the state-of-the-art LLMs make mistakes in directly generating netlists, they have been trained with extensive AMS circuit knowledge in human languages. As shown in Fig. \ref{fig:wrong_right}, they can provide promising design strategies at a circuit component level. Transforming these strategies to full netlists then becomes a much more systematic task. This paper addresses the critical limitations of LLMs in AMS netlist generation with a data-driven approach. Fig. \ref{fig:overall} presents the proposed AMSnet-KG, a comprehensive knowledge graph dataset containing netlists, architectural description (e.g. uses cascode current mirror, provides high DM gain), pin functions (e.g. $V_{in+}$, $I_{ref}$), and expert insights (e.g. parameter sharing between devices, operating point requirements). The architectural descriptions are summarized into \textit{global annotations}, and are used to index circuits and match LLM-generated design strategies. The netlists are used to circumvent LLMs' inaccuracy in generating topology. The pin functions and expert insights are summarized into \textit{local annotations}, and are used to guide assembly and automatic sizing. Building upon AMSnet-KG, we propose AMSgen, an automatic AMS circuit generation framework. AMSgen aims at generating fully sized netlists from performance specifications. Given design specifications, AMSgen starts by generating high-level design strategies, describing the circuit architecture using a number of circuit components. Matching circuit components are efficiently retrieved from AMSnet-KG and assembled into complete a topology for simulation. Afterward, guided by the local annotations to reduce the search space, AMSgen automatically optimizes transistor sizing through Bayesian optimization (BO). Finally, if the generated design does not meet performance specifications, AMSgen formats the current design and achieved performance into additional fewshot examples, and returns to the LLM-based strategy generation step. The flow iteratively generates new designs until the performance goals are achieved.

The key contributions of this paper are as follows:
\begin{itemize} 
    \item We construct AMSnet-KG, a comprehensive dataset thoroughly annotating AMS netlists with architectural description, pin functions, and expert insights. This labeled dataset is then arranged into a knowledge graph facilitating effective retrieval. The raw data for AMSnet-KG sources from existing literature, and are transformed into the proposed format.
    \item We propose AMSgen, an automatic AMS circuit generation framework that produces fully sized circuits from performance specifications. Steps include high-level design strategy generation (e.g., generating circuit architecture with circuit components), matched components and testbench retrieval from AMSnet-KG for automated assembly to obtain completed circuit topology, then device sizing guided by circuit annotations. Such architecture/topology can be regenerated if performance specifications with area constraints are not met.
    \item We apply AMSnet-KG through AMSgen and design circuits under a 28nm technology, achieving various desired performance specifications with minimal human effort.
\end{itemize}

The rest of the paper is organized as follows. Section \ref{sec:related_work} summarizes related work of LLMs for EDA, analog circuit datasets, and AI based AMS topology design. Section \ref{sec:amsnet_construction} provides the details of the AMSnet-KG dataset construction. Section \ref{sec:amsgen} describes the AMSgen, the automatic AMS circuit design flow based on AMSnet-KG and KG-RAG techniques. Finally, Section \ref{sec:experiments} presents two case studies, from required performance specifications to fully sized transistor-level netlists. Section \ref{sec:conclusion} concludes this paper and discusses future research directions.

\section{Related Work}
\label{sec:related_work}

\subsection{Analog Circuit Datset}
Due to the scarcity of open-source AMS circuit data, LLMs cannot achieve satisfactory AMS design results. AMSnet collects many circuit schematics from textbooks and academic papers and generates corresponding netlists\cite{tao2024amsnet}. AICircuit provides nine different types of AMS circuits with circuit schematics and a large number of simulation results for sizing\cite{mehradfar2024aicircuit}. \cite{ads} contains five types of AMS circuits, including schematic images, netlists, and testbenches. ALIGN provides a comprehensive collection of AMS circuits along with some well-sized parameters\cite{align_public}. AncstrGNN Benchmark collects many analog circuit netlists without schematic images\cite{chen2021universal}. Previous open-source datasets mostly contained only circuit schematic and netlists, which were not directly usable for LLMs. In this work, we expand and annotate the AMSnet dataset, enabling smoother retrieval by LLM during design.

\subsection{Machine Learning for AMS Circuit Design}
The automatic design of AMS circuits primarily includes topology synthesis and transistor sizing. In the era before LLMs, traditional machine learning (ML) methods for automatically designing AMS circuits were primarily divided into two types: 1) evolutionary algorithms (EA) based methods and 2) graph-based methods. The EA-based generation of circuit topologies, such as genetic algorithms and evolutionary algorithms, typically involves encoding circuit topologies into forms such as trees \cite{mcconaghy2011trustworthy} and graphs \cite{ak2014evolutionary}. New circuit topology can be generated through genetic operations, allowing for the evolutionary breeding of circuit designs. In graph-based methods, a comprehensive building block library is established, and circuit topology generation is conducted through the construction of graphs. The predefined building block library includes components such as resistors, capacitors, or any subcircuit (like current mirrors, etc.), and these elements are combined using methods based on reinforcement learning or similar techniques. This approach allows for efficient and intelligent assembly of complex circuit designs.

LLMs have demonstrated remarkable capabilities in numerous EDA tasks, such as Verilog and RTL code generation\cite{verigen}. Recent efforts have also explored using LLMs to directly generate circuit topology \cite{Artisan, lai2024analogcoder}. However, due to the scarcity of AMS circuit topology data in pretrain datasets, these efforts have not been very effective. Artisian\cite{Artisan} trained an LLM for the automatic behavior-level design of OPAMPs. However, the OPAMP-specific knowledge used to train their LLM cannot be smoothly generalized to other types of circuits, such as comparators and LDOs. Furthermore, it models circuits at a behavior level, resulting in higher transistor counts than a standard design. AnalogCoder \cite{lai2024analogcoder} did not retrain the model but used carefully designed prompts to leverage the LLM's capability to generate netlists. Due to the precise requirements of AMS circuits, multiple attempts and iterations are needed to achieve accurate results if possible. In this work, by using LLM for design strategy (e.g., topology category) selection rather than direct netlist generation, our design process provides more robustness.


\subsection{Sizing}
The goal of analog integrated circuit sizing is to determine the parameters of circuit components, such as transistors, within a given topology to meet design objectives. However, the substantial intra-class and inter-class variations, coupled with varying design goals across different circuit types, render the sizing process for analog integrated circuits particularly challenging. Researchers have extensively studied the automation of this process, primarily categorizing the methods into two types: 1) knowledge-based and 2) optimization-based methods. For knowledge-based methods such as \cite{horta2002analogue}, circuit designers create predefined schemes and equations to calculate transistor sizes. However, deriving design schemes for every existing circuit topology is very time-consuming, and also requires continuous maintenance to keep up with the latest process technologies. Notably, LADAC \cite{liu2024ladac} designed the first LLM agent for analog circuit sizing. By constructing an expert knowledge base, integrating ICL and CoT for decision-making, LADAC successfully derived transistor parameters for multiple circuit topologies, and satisfied their respective design specifications. Optimization-based methods treat the performance of circuits as a black-box function and use heuristic algorithms \cite{phelps2000anaconda,vural2012analog,liu2009analog,saǧlican2023moea,EA,gmid,MOEA} or surrogate model approaches for optimization \cite{xing2024kato,he2020efficient,gu2024tss,D3PBO,RL}. The use of Gaussian processes as surrogate models in optimization methods has been widely studied. ADO-LLM\cite{yin2024ado} combines an LLM with Bayesian Optimization (BO), using both the LLM and the acquisition function to determine the next sampling point. Yet, they did not discuss how to incorporate PDK technology nodes into their workflow, despite that PDKs play an important role in the sizing process. Previous optimization-based methods included expert insights to reduce search space\cite{nns}, but did not provide a systematic method to store and use these insights. Therefore, we propose an universal format in this work to store expert insights for transistor sizing.

\section{AMSnet-KG Construction}
\label{sec:amsnet_construction}
This section discusses the methods used to construct the AMSnet-KG dataset. We collect raw schematics and descriptions from literature sources, and label schematics in pages as well as components in schematics. We then propose a connectivity detection algorithm to create the netlists. Finally, we build the knowledge graph (KG) AMSnet-KG with collected schematics, netlists, and detailed annotations, to be defined below.

\begin{figure}[h!]
    \centering
    \includegraphics[width=\textwidth]{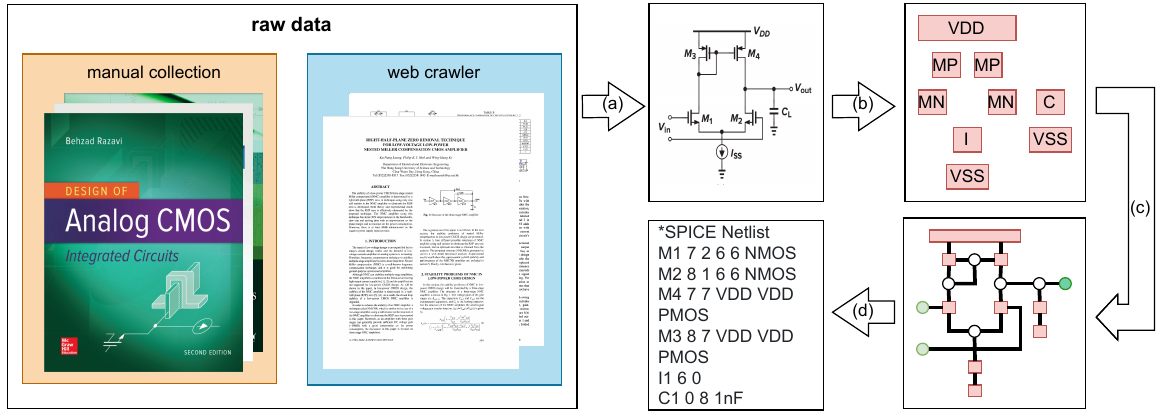}
    \caption{AMSnet construction pipeline. We collect a number of textbooks and papers, from which extract circuit schematic images are extracted. Then the flow detects circuit components, symbols, and nets from images, and generates full netlist.
    \\(a) schematic extraction and filtering, (b) component identification, (c) net identification, (d) netlist generation}
    \label{fig:pipeline}
\end{figure}

\begin{figure*}[h!t]
  \centering
  \begin{minipage}[b]{0.85\textwidth}
    \centering
    \includegraphics[width=\textwidth]{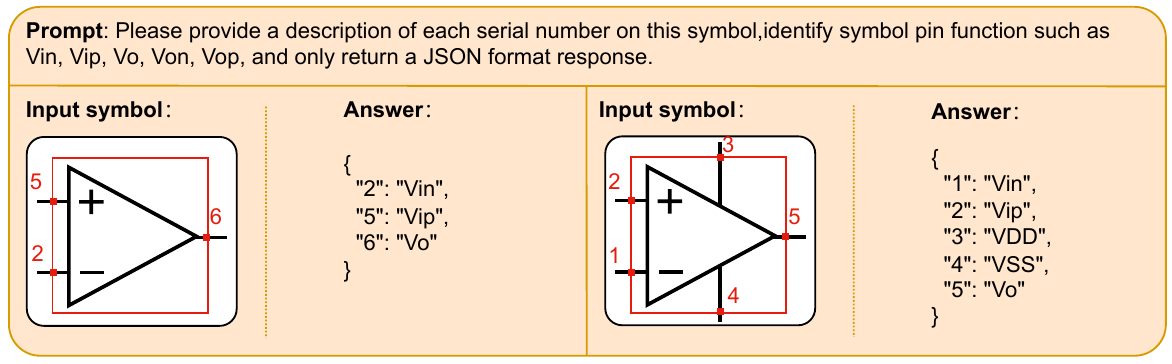}
    \caption{An example of symbol pin identification by GPT4}
    \label{fig:askGPT}
  \end{minipage}
\end{figure*}


\subsection{Data Collection}
As shown in Fig. \ref{fig:pipeline}, we first collect large quantities of raw data from textbooks and academic papers. These materials are rich in circuit schematics and verbal description, and provide sufficient foundation to form our <schematic, netlist, annotation> dataset. To reduce the cost of manually extracting circuit schematics, we employ a semi-supervised learning approach. We annotate bounding boxes over schematics on a subset of page images, and train an object detection model with the labeled data. This model is then used to identify and extract all circuit schematics from the remaining pages.

Since most literature print their schematics, individual components are generally printed in a very uniform fashion, which allows us to perform template matching. Instead of manually annotating bounding boxes on components within schematics to form a training set as we did with literature pages, we only need to annotate a single copy of each component type. For example, after annotating a PMOS transistor, the region of interest (RoI) within the bounding box can be template-matched against the other schematics to quickly identify all other PMOS transistors. This drastically reduces the amount of effort required for manual labeling. It is important to note that the orientation information will be used in later steps to determine component connectivity. Therefore, we require a label for each orientation of each component class.

Despite being more cost-effective than manual labeling, template matching remains time-consuming compared to deep-learning models. We therefore continue with the semi-supervised approach, where we train a model using data from template matching, and apply the model to label the remaining schematics. Unlike components, subcircuit symbols are not always printed uniformly, such as the OPAMPs in Fig. \ref{fig:askGPT}. Hence, they require large numbers of manual labeling to train a similar detection model. We manually evaluated the \textit{recall} metric of the component and symbol detection results, and the overall recall was 97\%. At this point, we have the location labels and class labels for all components in our schematics.

\subsection{Netlist Generation from Schematic Image}
After identifying the components, we can label the net connections. The current version relies on two assumptions: 1) all the wires are represented by solid lines on the schematic diagram, and 2) without a junction, two intersecting wires are not considered connected. These assumptions enable us to implement the net detection algorithm as follows.
\begin{figure}[ht]
  \centering
  \begin{minipage}[b]{0.35\textwidth}
    \centering
    \includegraphics[width=\textwidth]{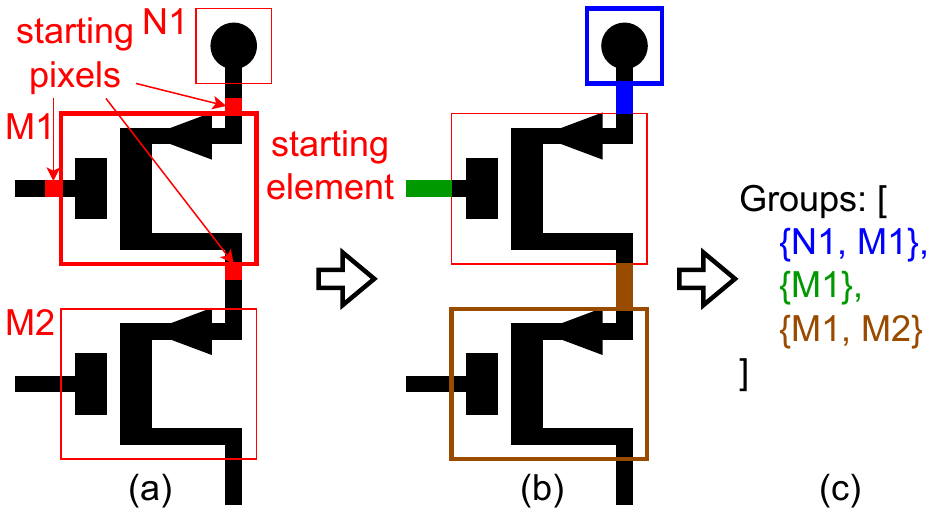}
    \caption{Grouping neighboring components for net detection}
    \label{fig:bfs_basic}
  \end{minipage}
  \hfill
  \begin{minipage}[b]{0.35\textwidth}
    \centering
    \includegraphics[width=\textwidth]{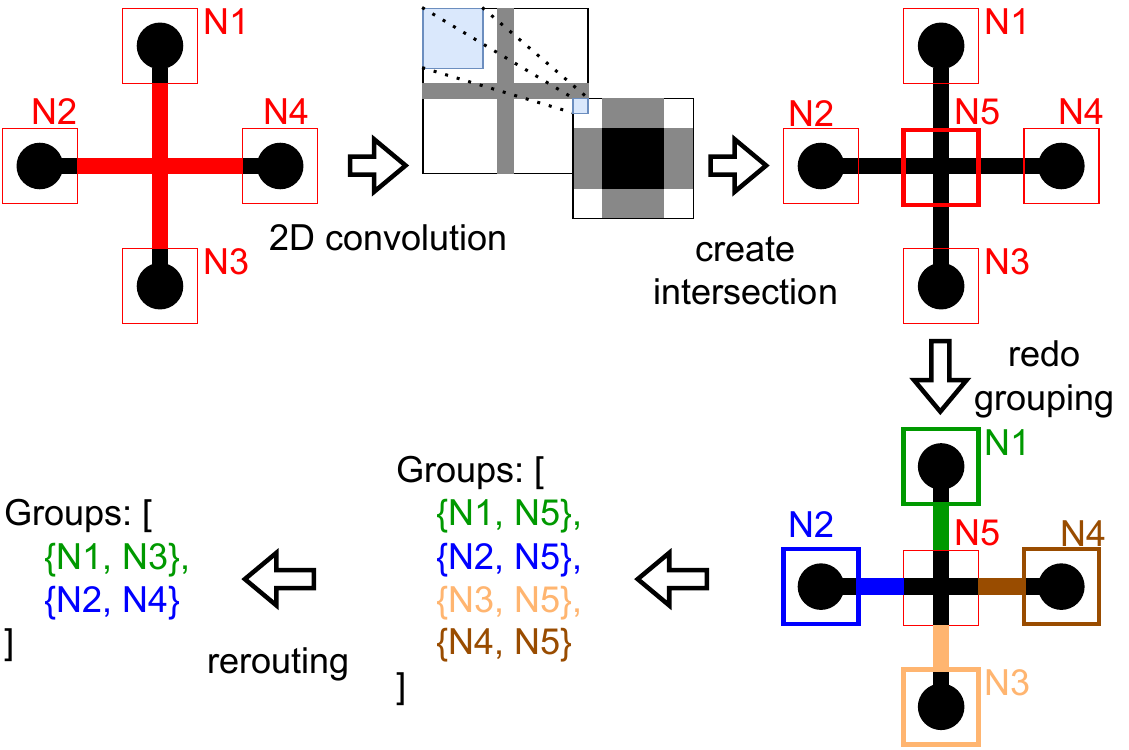}
    \caption{Resolving intersection cases in net detection}
    \label{fig:2d_convolution}
  \end{minipage}
  \hfill
  \begin{minipage}[b]{0.25\textwidth}
    \centering
    \includegraphics[width=\textwidth]{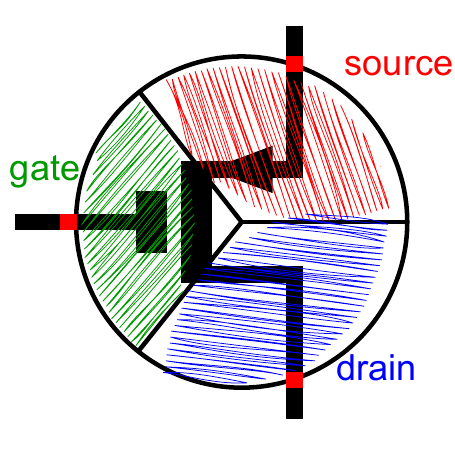}
    \caption{An example of component pin ordering}
    \label{fig:connection_ordering}
  \end{minipage}
\end{figure}

The first step is to group all neighboring components. Starting from each pixel on each bounding box, the algorithm expands into neighboring wire pixels in all directions, until it encounters other components. This step groups directly components into clusters, each representing a net. Fig. \ref{fig:bfs_basic} shows an example.



Each group has four possible cases: 1) The group contains only the starting component, as shown in the green group in Fig. \ref{fig:bfs_basic}. In this case, no connectivity has been detected. 2) The group contains exactly two components, as depicted by the blue and brown groups in Fig. \ref{fig:bfs_basic}; here, the two components are connected. 3) The group has an odd number of components (more than two). This scenario likely indicates that a junction has been omitted. However, the algorithm cannot determine which subset of the group is connected and which is intersecting. Therefore, we flag the entire schematic as an exception; manual attention is required to correct it before it can be analyzed again.


For the last case, there are an even number (more than 2) of components in the group, as illustrated in Fig. \ref{fig:2d_convolution}. We assume the schematic has not omitted any junctions; thus, the intersecting wires are not connected. To address this, we locate the intersections by applying a 2D convolution to the searched wiring. Given that the area around the intersection typically contains a higher density of wire pixels, we identify the indices with the maximum values as the intersection point and add it to our labeled components. Later, the algorithm eliminates the four-component cluster by repeating the grouping process, connecting each of the four components to the intersection. In the case of more than four components in the group, each iteration reduces the group size by two until each group is left with two components eventually. It is important to note that line weights and layout may influence this step, making the dimension of the convolution kernel a tunable parameter. We reroute opposite connections to each other, and then delete the intersection to finalize the process.


After all the nets are identified, the SPICE netlist format for some components requires the correct order of connections. For example, the connections to four-terminal MOSFETs must follow the order of drain, gate, source, and body / substrate. For components, the bounding boxes are labeled with orientation, and the algorithm can determine a range of angles for each connection, as shown in Fig. \ref{fig:connection_ordering}. For symbols, as shown in Fig. \ref{fig:askGPT}, we mark a number on each pin, and use MLLM to determine their function. The quality of the net labeling process was manually verified, which arrived at an accuracy of 96\%. Erroneous results are manually corrected to ensure data quality.

At this point, the algorithm identifies circuit components, symbols, and nets in the given schematic diagrams. Using this information, we are able to generate the unsized netlist.

\begin{figure}[h!t]
    \centering
    \includegraphics[width=\textwidth]{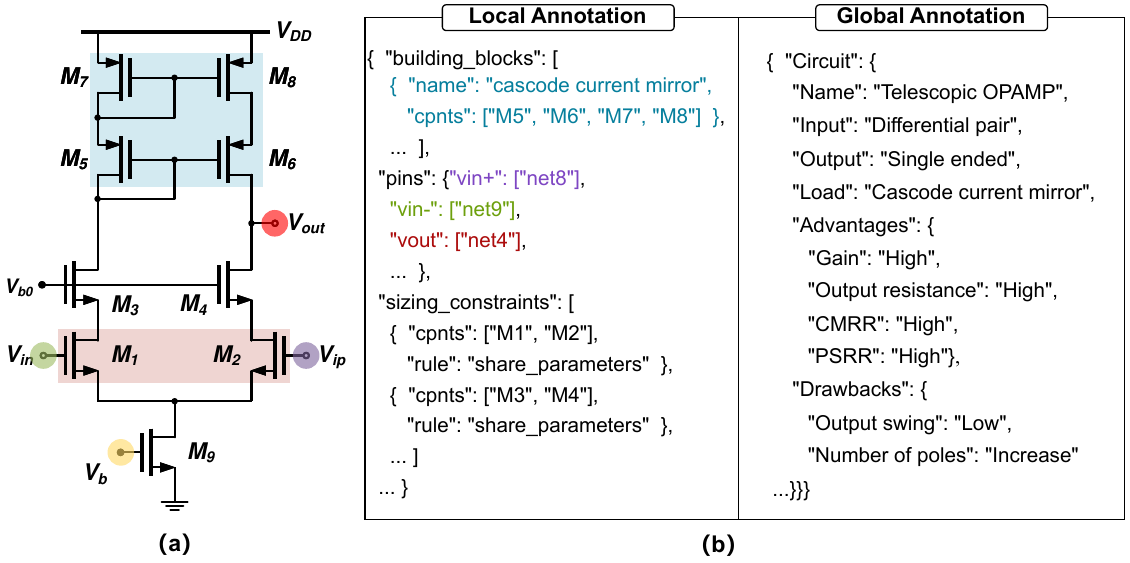}
    \caption{(a) An example of component labeling, (b) corresponding annotations.}
    \label{fig:json}
\end{figure}

\subsection{Circuit Annotation}
In addition to schematics and netlists, we also include annotations to enrich circuit data. We define two types of annotations: \textit{local} and \textit{global}, where local annotations describe one or more component(s) of the circuit, while global annotations describe the entire architecture.

The purpose of local annotations is to guide \textit{usage} of the current circuit, and is hand-labeled in this work. Some examples include identifying nets as inputs, biases, outputs, or identifying groups of components as building blocks such as current mirror or differential pair. An important type of local annotation is to label sizing constraints, marking certain sets of components to follow constraints such as symmetry in lengths and widths. These constraints are usually expert insights gained from their design experience, and play an important role in the sizing process to reduce search space and computation costs. Net labels, on the other hand, are used to connect pins of circuits to testbenches.

The purpose of global annotations is to guide topology selection, to use a circuit as opposed to other circuits. It is common that each individual circuit does not have a generally agreed name to identify them. Instead, they are described by circuit components or remembered by their specialties. An engineer may describe an OPAMP with its stages, or decide to use it for its exceptional performance in a specific area. To enable this information retrieval process, we label each circuit with its architectural descriptions such as performance specialties and circuit subcomponents. Specifically, we exploit the fact that the source literature where we first obtained each schematic, would most likely include some verbal description in the context. We use this information in addition to the schematic image and invoke MLLM to summarize qualitative description into key-value pairs, such as \{gain: high\} or \{load: current mirror\}.

We take one step further and produce JSON files with LLM to introduce more structure into our annotation data. Fig. \ref{fig:json} (a) illustrates annotation on a circuit, while Fig. \ref{fig:json} (b) presents the local and global annotation files for the given circuit. Since individual JSON files may share the same annotation, we simplify the data retrieval process and reduce data volume by merging equivalent annotations to form a knowledge graph in the following section.

\subsection{Knowledge Graph Construction}
A knowledge graph (KG) is a structured form of knowledge representation that expresses the relationships between entities. In a knowledge graph, nodes typically represent entities, and edges signify various semantic relationships between these entities. The basic unit of composition is the ``entity-relation-entity'' \textit{relation query triplet}, where entities are interconnected through relations, forming a graph-like data structure. In this work, we collect all schematics, netlists, and annotation data to create the AMSnet-KG dataset.

\begin{figure}
    \centering
    \includegraphics[width=\textwidth]{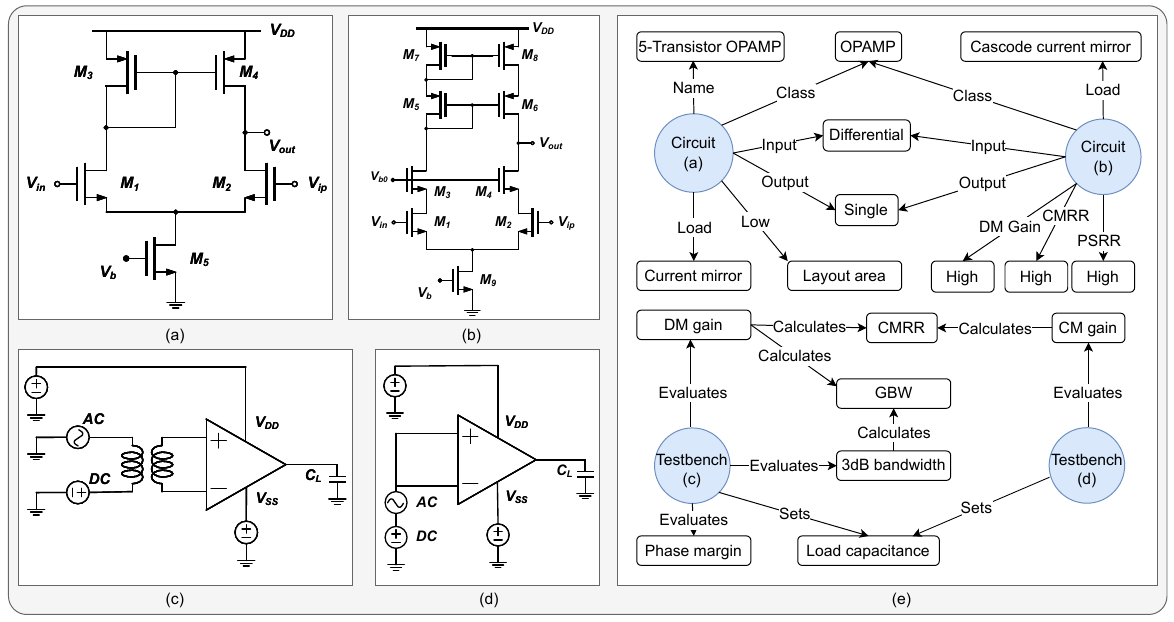}
    \caption{(a) 5-Transistor OPAMP, (b) Telescpoic cascode OPAMP, (c) Testbench for DM gain, (d) Testbench for common mode gain, (e) Corresponding knowledge graph diagram.}
    \label{fig:knowledge_graph}
\end{figure}

AMSnet-KG defines an entity as either a string or a circuit. The annotation keys and values are simply represented as strings, while the circuit object contains a schematic, a netlist, a set of image attributes such as bounding boxes and net marks, and all local annotations. Relations, on the other hand, only serve to connect entities and are therefore also simply strings. Fig. \ref{fig:knowledge_graph} illustrates a small scale knowledge graph created from two circuits and two testbenches.



\subsection{Dataset Summary}


To summarize, AMSnet currently contains 894 circuits with schematic diagrams, component bounding boxes, and netlists. AMSnet-KG selects application-specific circuits and further include local and global annotations. It currently contains OPAMPs, comparators, bandgaps, LDOs, and ADCs, as well as related testbenches.


\section{AMSgen: AMSnet-KG Driven Automated Circuit Generation}
\label{sec:amsgen}
\begin{figure}
    \centering
    \includegraphics[width=\textwidth]{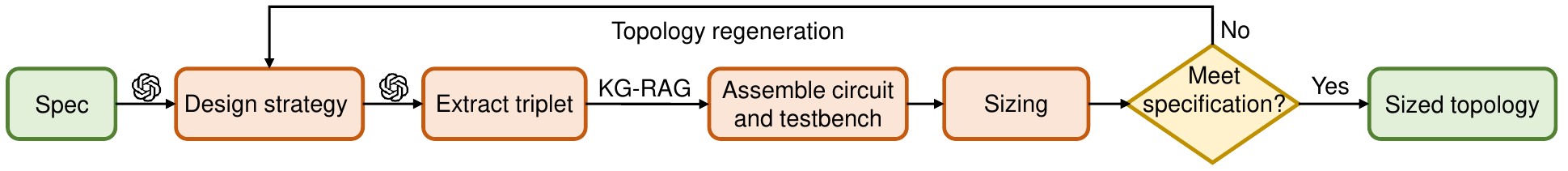}
    \caption{Pipeline of AMS circuit design from performance specifications to fully sized transistor-level netlist}
    \label{fig:select_pip}
\end{figure}

The data support from AMSnet-KG enables us to fully automate AMS circuit design, from performance specification to fully sized netlists. In this section, we explore our LLM-assisted, data-driven design pipeline, as shown in Fig. \ref{fig:select_pip}, which autonomously selects circuit topology and simulation testbenches in section \ref{sec:topology_selection}, and optimizes transistor sizing in section \ref{sec:sizing}. In the case where the initial topology fails to reach performance goals after sizing, we automatically regenerate the topology in section \ref{sec:topology_regeneration}.

\begin{figure}
    \centering
    \includegraphics[width=\textwidth]{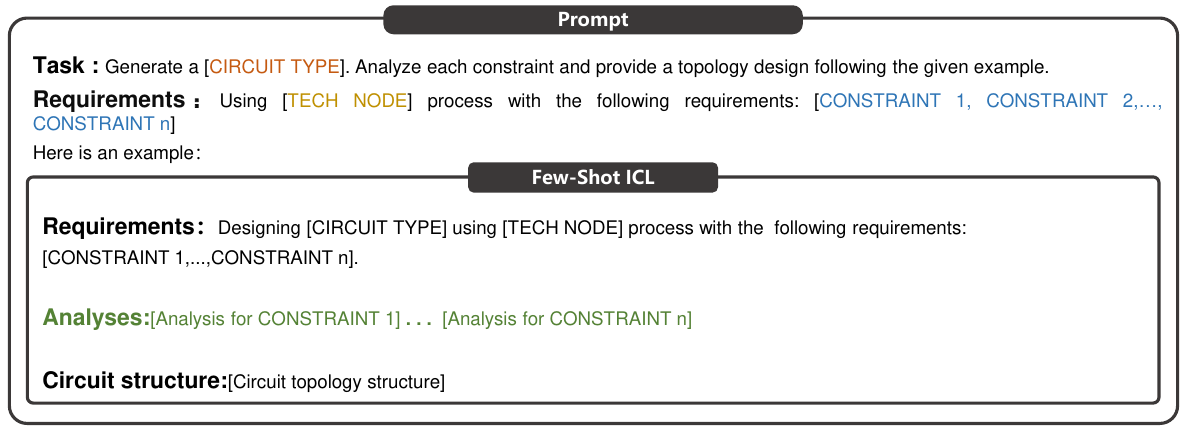}
    \caption{LLM conversation: from performance specification to design strategy (shortened, full version in the Appendix section \ref{appendix:llm_conversations}). The green text presents CoT prompt.}
    \label{fig:prompt_shot}
\end{figure}

\subsection{Topology Selection via KG-RAG}
\label{sec:topology_selection}
Traditionally, analog / mixed-signal (AMS) circuits are designed manually by engineers, starting with circuit topology and the testbench selection. At the outset, the engineer considers a set of desired performance specifications. Drawing from their experience or supplementary materials, they identify circuit topologies that meet the design goals and select testbenches to evaluate each performance metric. However, this process is highly labor-intensive, time-consuming, and experience-driven, making it difficult to automate or abstract into code such as RTL. With the LLM technology available today in addition to our AMSnet-KG data support, we are able to fully automate this process and achieve the same goals. Given a set of performance metrics, we obtain a design strategy from the LLM which outlines the circuit architecture as a set of circuit components. Then, we select the corresponding circuits and testbenches from AMSnet-KG, before assembling them for simulation.

We design a comprehensive prompt engineering framework, incorporating in-context learning (ICL) \cite{icl1,icl2} and chain of thought (CoT) \cite{cot}, for generating circuit topology design strategy, a short verion is shown in Fig. \ref{fig:prompt_shot}, while full versions are shown in the Appendix section \ref{appendix:llm_conversations}. The state-of-art LLMs possess sufficient knowledge to provide a design strategy for desired circuits. However, they answer questions in natural text for human users, which may be difficult to extract and use programmably. To standardize response format, we use an ICL strategy with an example(s), encouraging the LLM to emulate this output format, which facilitates the further application of the response. We also employ CoT prompting methods as they proves to effectively enhance the performance of LLMs. Specifically, the model is instructed by example to first analyze the requirements in the specifications qualitatively, explain a thought process, before providing the actual circuit.

\begin{figure}
    \centering
    \includegraphics[width=\textwidth]{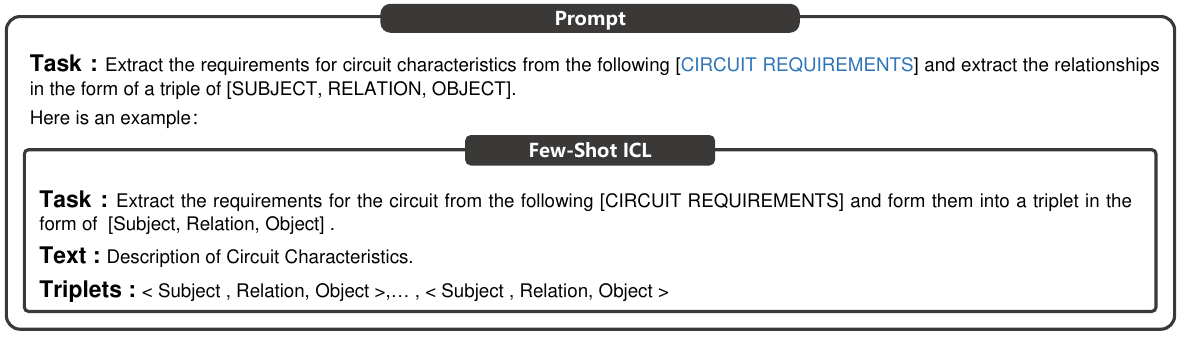}
    \caption{LLM conversation: from circuit characteristics to extract triplets (shortened, full version in the Appendix section \ref{appendix:llm_conversations})}
    \label{fig:prompt_shot_02}
\end{figure}

After receiving the LLM's analysis and response regarding the design specifications, we format the LLM's responses into relation query triplets to facilitate the retrieval of relevant circuits from AMSnet-KG, as illustrated in Fig. \ref{fig:prompt_shot_02}. Similarly, to standardize the output format and reduce mismatches during the retrieval of triplets in the knowledge graph, we design few-shot examples and input them to the LLM along with the intended question. After obtaining the triplets, we perform searches within the knowledge graph using matching queries, such as those formulated in the Cypher language for use in Neo4j as follows: 

\texttt{MATCH (node:circuit\{input:'Differential'\}) RETURN node ORDER BY ...}

Notably, this step serves a similar purpose to the embed-and-rerank step in standard RAG. With the retrieved circuits and testbenches, we make use of net functions preserved in local annotations, and assemble the full circuits by matching net names with each other (i.e. first stage ``Vout'' to second stage ``Vin''). A detailed example is shown in section \ref{sec:opamp_topology_selection}. This builds a fully functional netlist for simulation, and prepares us for the upcoming sizing step.

\subsection{Constraint-Augmented Sizing}
\label{sec:sizing}

Now that the circuit and testbenches have been fixed, we optimize component parameters to obtain maximal performance achievable with the given topology. In this section, we formulate sizing as a black-box optimization problem and resolve it through Bayesian optimization (BO) with local annotations retrieved from AMSnet-KG. The details of BO are included in Appendix section \ref{sec:bo}. 

\subsubsection{Problem Formulation}
The standard constrained optimization problem is shown in Equation (\ref{eq:opt_problem}):
\begin{equation}
\label{eq:opt_problem}
\begin{aligned}
& \text{maximize} \quad FoM(x) \\
\text{s.t.} \quad & g_j(x) \leq 0, \quad \forall j \in \{1, \ldots, p\} \\
\end{aligned}
\end{equation}
where figure of merits (FoM) is the objective function, $g_j(x)$ is the $j$-th performance constraint.
Following \cite{xing2024kato,wang2020gcn}, we define FoM as weighted square sum of the normalized performance value as shown in (\ref{eq:fom}):
 \begin{equation}
\label{eq:fom}
\begin{aligned}
FoM = \sum_{i=0}^{N} w_i \times \frac{\min(f_i(x), f_i^\text{bound}) - f_i^{\min}}{f_i^{\max} - f_i^{\min}}
\end{aligned}
\end{equation}
where $f_i(x)$ is $i$-th simulated performance, $f_i^\text{bound}$ is performance limit which we do not get additional merit after it has been reached, $f_i^{\min}$ and $f_i^{\max}$ are normalization factors obtained through an initial set of random samples, and $w_i$ is the term weight to control the importance of each performance metric.

BO is a powerful black-box optimization method to search for the optimal transistor sizing\cite{xing2024kato}, by balancing between exploration and exploitation. BO uses a series of initial data to define a surrogate model (typically a Gaussian process), selects the next sampling point by maximizing the acquisition function, and updates the surrogate model using the new sampled data. Exhaustively searching parameters for all transistors is inefficient, and therefore we use expert insights as \textit{constraints} to the search space. The local annotations stored in AMSnet-KG can effectively reduce the parameter search space during exploration and exploitation. For instance, with the annotation that a set of transistor parameters should maintain equality or a certain ratio, we reduce the number of parameters accordingly. The algorithm flow for constraint-augmented sizing is as follows:

\begin{algorithm}
    \SetAlgoLined
    \textbf{inputs:} initial data size $N_{init}$, number of iterations $N_{iter}$, constraint-augmented parameter design space $S_{param}$\;
    $x_{init} \gets$ sample($S_{param}$, $N_{init}$)\; 
    $y_{init} \gets$ performance output from simulation with $x_{init}$\; 
    fit $\mathcal{GP}(x_{init}, y_{init})$\;
    \For{$t \gets 0$ \KwTo $N_{iter}$}{
        \text{// select next points for simulation via acquisition function $\alpha(x)$}\\
        $x_{next} \gets \arg \max_{x} \alpha(x)$\; 
        $y_{next} \gets$ performance output from simulation with $x_{next}$\; 
        fit $\mathcal{GP}(x_{next}, y_{next})$\;
    }
    \textbf{return} \text{optimal sizing $x$*}\;
    \caption{Constraint-Augmented Sizing for Analog Circuit Design}
    \label{alg:bo}
\end{algorithm}



\subsection{Topology Regeneration}
\label{sec:topology_regeneration}
In the case where an initial topology design does not satisfy the desired metrics after a set amount of sizing effort (or it costs too much circuit area for the desired results), we consider this a topology issue and begin regeneration. The topology regeneration process is very similar to the initial topology design process. The only difference is that instead of using a different circuit application in the fewshot examples, here we use actual performance specifications obtained from the previous design(s) to give the LLM additional quantitative knowledge. 

With the previous design strategy proposed by the LLM and the optimal performance during simulation, we reverse the cause-effect and pretend that we initially \textit{wanted} to achieve the performance metrics, and then \textit{correctly} obtained the previous design. For example, suppose we desire an OPAMP with a DM gain of 60 dB, the LLM initially proposed a design strategy to retrieve circuit components and assemble circuit \textit{A}. Then we go through the process outlined in section \ref{sec:sizing}, and finally ended with a fully sized design that only achieves a DM gain of 40 dB. We would now inject a fewshot example where we \textit{desire a design that achieves a DM gain of 40 dB, and then obtained circuit A}; and then prompt the LLM for a new design that achieves 60 dB instead. The LLM would then understand to propose a topology with higher expected gain.

This approach is especially effective since initially the state-of-art LLM does not possess sufficient quantitative knowledge. It is unable to accurately infer the performance of a specific circuit topology on a specific technology node, and therefore generally begins with a balanced design that has no particular weakness. However, it does possess qualitative and comparative knowledge between different topology, as they generally hold across all technology nodes. In this example where we require a higher gain, it is often able to correctly suggest a new topology that specializes in higher gain. With the new topology, we simply replay netlist retrieval, assembly, and sizing for the same set of performance specifications. This process is repeated until we either achieve the intended performance goals or exhaust all possible topologies, where we either finalize our design or label the input as impossible to achieve.

\section{Experiments}
\label{sec:experiments}
\subsection{Experiment Setup}

We use YOLO-V8 for object detection during AMSnet-KG construction, Neo4j for knowledge graph implementation, and GPT-4 for various inquiries throughout dataset construction and circuit design. Our experiments are performed using a 28nm technology from SMIC, which restricts our transistor length, width, and number of fingers to [30nm, 1$\mu$m], [100nm, 3$\mu$m], and [1, 100] respectively. We use Bayesian optimization (BO) implemented by the Optuna \cite{akiba2019optuna} library, and limit our total simulation count to 2000.

\subsection{Case Study of OPAMP Topology Design}
\subsubsection{Topology selection}\ 
\label{sec:opamp_topology_selection}

Initially, we request an OPAMP design to achieve the performance goals in Equation \ref{eq:opamp_spec}. 
\begin{equation}
\label{eq:opamp_spec}
\begin{aligned}
\text{Gain} > 80\,\text{dB}, \quad \text{CMRR} > 80\,\text{dB}, \quad \text{PSRR} > 80\,\text{dB}, \quad \text{GBW} > 10\,\text{MHz}, \quad \text{PM} > 60^\circ, \quad C_L = 100\,\text{pF}
\end{aligned}
\end{equation}


In order to keep any incorrect numerical information from the in-context learning, we use a different type of circuit to offer the reasoning format and chain of thought. Specifically, we use a comparator design flow as a fewshot example. Fig. \ref{fig:prompt_01_example} and Fig. \ref{fig:prompt_01_response} in the Appendix section \ref{appendix:opamp_initial_design} presents the full conversation. In this case, the LLM responds with the design strategy of a two-stage OPAMP, where the first stage is a differential amplifier with a current mirror load, the second stage is a common-source amplifier, and a Miller compensation is to be added between the outputs of the first and second stage.

\begin{figure*}
    \centering
    \includegraphics[width=\textwidth]{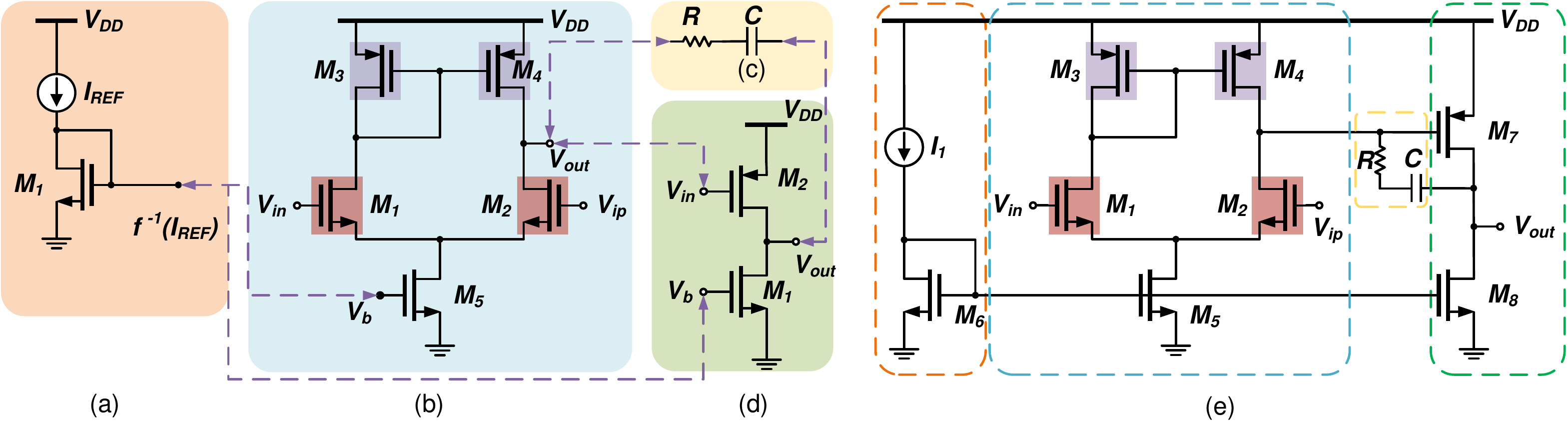}

    \caption{Visualization of the retrieved circuit components followed by the assembled result, the purple arrows illustrate circuit component -level connectivity. (a) bias circuitry, (b) first stage, (c) Miller compensation, (d) second stage, (e) assembled 2-stage OPAMP, symmetric transistors identified from local annotations are highlighted in the same color.}
    \label{fig:full_circuit}
\end{figure*}

We then convert this design strategy to a set of relation query triplets, Fig. \ref{fig:prompt_02_example} and Fig. \ref{fig:prompt_02_response} in the Appendix section \ref{appendix:opamp_initial_design} presents the full conversation. The triplets are then used to query AMSnet-KG and retrieve circuits. For example, <{}, input, differential input pair>, <{}, load, PMOS current mirror> retrieves a single stage 5-transistor OPAMP. Similarly, we retrieve the common source amplifier, the bias circuitry with a current source and an NMOS transistor, and the sequentially connected resistor-capacitor pair for Miller compensation. Using the local annotations on each circuit component, we connect the output from the first stage to the input of the second stage, the bias voltage of both stages to the bias circuitry, and a Miller compensation to the output of the first stage and the output of the second stage. The fully assembled 2-stage OPAMP is shown in Fig. \ref{fig:full_circuit}.

Similarly, we convert the required performance metrics to another set of relation triplets. Using this information, we retrieve three testbench circuits to test the desired metrics, including DM gain, CM gain, PS gain. The testbench circuits are connected to our 2-stage OPAMP by matching net annotations similar to above. We setup SPICE simulation with the Spectre circuit simulator by Cadence, and ensure that the invocation can be done programmably to support the acquisition function used by BO in the next step. At this point, we are ready to perform simulation and collect performance metrics.


\subsubsection{Sizing}
The given specifications forms the FoM in Equation \ref{eq:opamp_fom}, where $f_{gain}(x)$, $f_{cmrr}(x)$, $f_{psrr}(x)$, $f_{gbw}(x)$, and $f_{pm}(x)$ each represent the simulated gain, CMRR, PSRR, gain-bandwidth product, and phase margin. Each $f^{min}$ and $f^{max}$ represent the corresponding minimum and maximum metric for normalization purposes, and are observed within an initial set of 100 simulations. Note that since the GBW distribution is exponential, we first take the decimal log of each value before using it for FoM to prevent it from overtaking other values. As for phase margin, instead of requiring a high value, we actually want the value to be as close to 60 degrees as possible, therefore we introduce its distance away from 60 degrees as a penalty.

\begin{equation}
\label{eq:opamp_fom}
\begin{aligned}
FoM =& \frac{\min(f_{gain}(x), 80) - f_{gain}^{\min}}{f_{gain}^{\max} - f_{gain}^{\min}} + \frac{\min(f_{cmrr}(x), 80) - f_{cmrr}^{\min}}{f_{cmrr}^{\max} - f_{cmrr}^{\min}} + \frac{\min(f_{psrr}(x), 80) - f_{psrr}^{\min}}{f_{psrr}^{\max} - f_{psrr}^{\min}} + \\
& \frac{\min(\lg(f_{gbw}(x)), 7) - \lg(f_{gbw}^{\min})}{\lg(f_{gbw}^{\max}) - \lg(f_{gbw}^{\min})} - \frac{|f_{pm}(x) - 60|}{f_{pm}^{\max} - f_{pm}^{\min}}
\end{aligned}
\end{equation}

As shown in Fig. \ref{fig:full_circuit}(e), our design consists of 5 NMOS transistors and 3 PMOS transistors, each of which is described by two parameters: gate length ($L$) and gate width ($W$). In addition, we have a current source, a capacitor, and a resistor, each described by a single parameter, for a total of 19 free parameters.

Insights from experienced engineers could effectively reduce the complexity of the sizing process, such as the parameter search space. We make use of the local annotations stored in AMSnet-KG, as shown in Fig. \ref{fig:full_circuit}. Transistors with the same highlight color share the same set of $L$ and $W$. This way, our effective number of free parameters reduces from 19 to 15, and is more manageable.

\begin{figure*}
    \centering
    \includegraphics[width=0.45\textwidth]{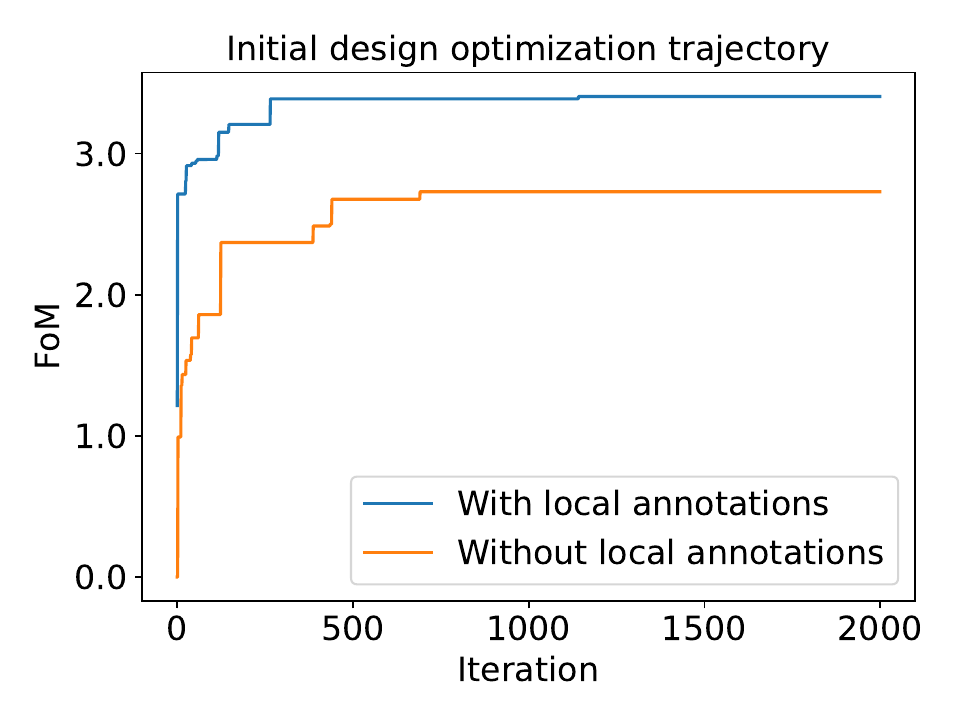}
    \caption{Bayesian optimization trajectory}
    \label{fig:bayesian_optimization_fom_opamp_initial}
\end{figure*}

Fig. \ref{fig:bayesian_optimization_fom_opamp_initial} shows the optimization trajectory for the initial design, comparing between \textit{constraint-augmented sizing} versus \textit{vanilla sizing}. We can see that with a smaller number of free parameters, convergence happens much earlier. A final design with the highest FoM is shown in Table \ref{tab:fom_result} rows (a) and (b) for the two attempts. We can see that attempt (b) significantly outperforms attempt (a), which demonstrates that the retrieved local annotations yields final performance improvements in addition to convergence time.

Unfortunately for this initial topology, the optimal sizing we obtained is not enough to satisfy our specification requirements. The gain value is only 66.21 dB, noticeably lower than the required 80 dB. Additionally, the CMRR and PSRR values also do not meet requirements. While it is possible that our sizing step did not reach the global optimal set of parameters for this topology, the more straightforward approach would be to regenerate our circuit topology.

\begin{table}[tb]
\centering
\scriptsize
\resizebox{\columnwidth}{!}{
\setlength{\tabcolsep}{1.2mm}
\begin{tabular}{l|l|l|l|l|l|l|l|l|l|l}
\toprule
    & Topology & BO constraint & FoM $\uparrow$ & DM gain (dB)$\uparrow$ & GBW (Hz)$\uparrow$ & CMRR (dB)$\uparrow$ & PSRR (dB)$\uparrow $& Phase margin ($^\circ$) \\ \midrule
Specification & - & - & - & 80.00 & $1.00 \times 10^7$ & 80.00 & 80.00 & 60.00 \\ \midrule
(a) & 2-stage & W/o local annotations & $2.73$ & 60.06 & $3.91 \times 10^7$ & 17.15 & 34.47 & 73.15 \\ \midrule
(b) & 2-stage & W/ local annotations  & $3.40$ & 66.21 & $\mathbf{3.19 \times 10^8}$ & 54.20 &  69.82 & 63.63 \\ \midrule
(c) & 2-stage high-gain & W/o local annotations & $2.55$ & 45.95 & $5.05 \times 10^7$ & 20.62 & 59.89 & 63.24 \\ \midrule
(d) & 2-stage high-gain & W/ local annotations  & $\mathbf{3.60}$ & \textbf{80.85} & $1.43 \times 10^7$ & \textbf{99.04} & \textbf{91.76} & \textbf{60.48} \\ \bottomrule
\end{tabular}
}
\caption{Designs optimized by Bayesian optimization}
\label{tab:fom_result}
\end{table}

\subsubsection{Topology Regeneration}
The regeneration process for LLM to propose another design strategy is very similar to the initial strategy proposal. Here, instead of using for performance specification requirements and design structure from other applications such as comparators in fewshot examples, we are now able to simply use the previous, unsuccessful design. This allows us to make a numerical comparison between the two circuits for a specification, and thus leverage the LLM's vast knowledge on comparative performance between circuits. As shown in Fig. \ref{fig:prompt_03_example} and Fig. \ref{fig:prompt_03_response} in the Appendix section \ref{appendix:opamp_regenerated_design}, we reverse the cause-effect order by pretending that we initially \textbf{wanted} a design to achieve a DM gain of 66.21 dB, gain-bandwidth product of 319MHz, etc. The analysis and design strategy would be the same as the previous circuit topology. Afterwards, by asking the same question, the LLM naturally gains a quantitative comparison between the existing performance specification in the fewshot example and the actual prompt. This resolves our previous issue where the LLM has insufficient knowledge regarding the specific technology node we are using.

\begin{figure*}
    \centering
    \includegraphics[width=\textwidth]{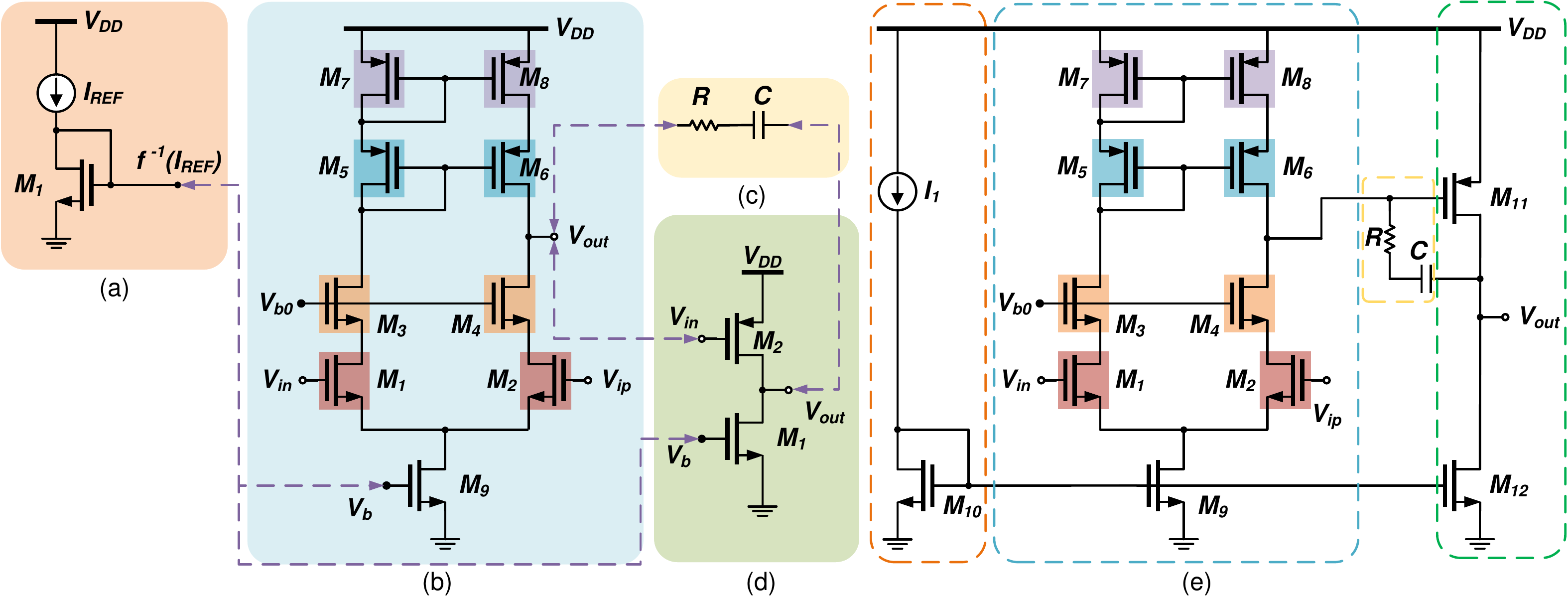}
    \caption{Visualization of the retrieved circuit components followed by the assembled result, the purple arrows illustrate circuit component -level connectivity. (a) bias circuitry, (b) first stage, (c) Miller compensation, (d) second stage, (e) regenerated 2-stage OPAMP, symmetric transistors are highlighted in the same color.}
    \label{fig:revised_circuit}
\end{figure*}

In our case, the DM gain value was too low, and therefore the natural reaction of an engineer would be to select a different topology that specializes in higher DM gain. The LLM follows a similar thought process and decides to update the first stage from the five-transistor design to a telescopic cascode, and keeps the rest of the design as-is. We then use this information in the same AMSnet-KG retrieval process as before, followed by an assembly process based on local annotations. The result circuit is illustrated in Fig. \ref{fig:revised_circuit}.

\begin{figure*}
    \centering
    \includegraphics[width=0.45\textwidth]{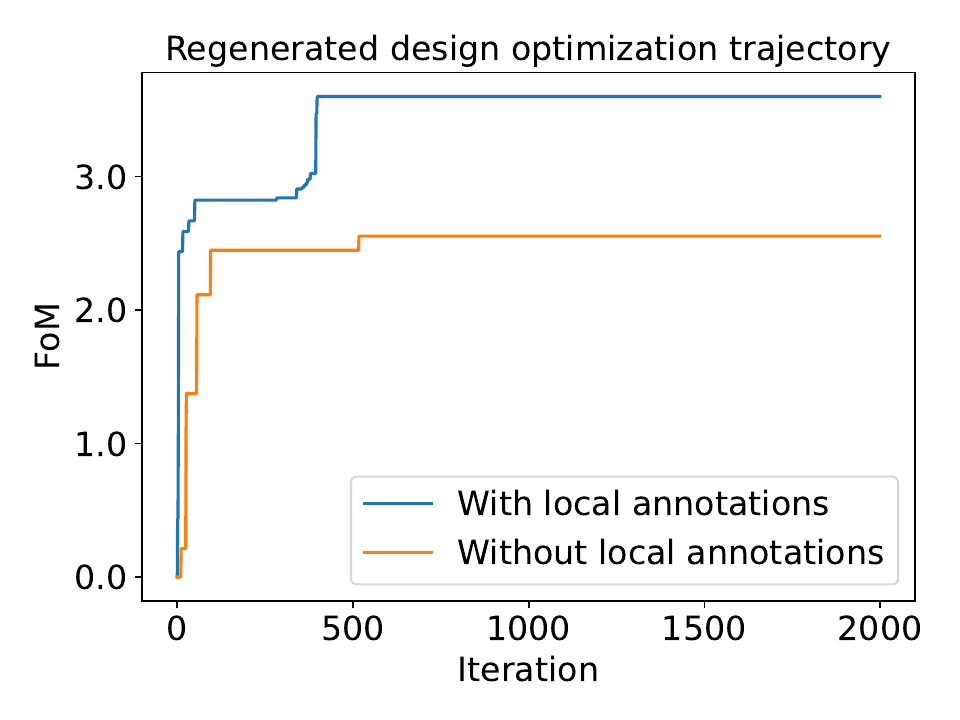}
    \caption{Bayesian optimization trajectory for regenerated design for higher DM gain}
    \label{fig:bayesian_optimization_fom_opamp_regenerated}
\end{figure*}

The sizing process is mostly the same as before, the only difference being that now there is a number of new transistors and a bias voltage. We first identify symmetry constraints, reduce the number of free parameters from 27 to 19. As we can see in Fig. \ref{fig:bayesian_optimization_fom_opamp_regenerated}, the optimization trajectory is very similar in shape as that of the previous topology. Quantitatively, the FoM score we achieve is much better at almost zero, indicating that we met every design goal. On the other hand, convergence happens later due to the higher parameter count and stricter specifications. Once again, the sizing attempt without constraining parameters takes much longer to gain FoM, and did not produce an acceptable design at the end of our 2000 iteration limit.

The final performance is shown in Table \ref{tab:fom_result} row (d), as we can see we met every design goal initially intended, proving our process successful.

\subsection{Case Study of Comparator Topology Design}
\subsubsection{Topology selection}
In order to evaluate the AMSgen workflow on different applications, we introduce another case study on comparators. Similar to generating OPAMPs, we begin by requesting a comparator design from LLMs, using the same PDK and constraints, and a comparator-specific set of performance specifications as follows:
\[ \text{Sampling frequency = 1GHz}, \quad \text{Offset voltage} < 100\,\text{uV}, \quad \text{Propagation delay} \leq 1\,\text{ns}, \quad \text{Power} \leq 100\,\text{uW} \]

The LLM first responds with the use of a latch comparator. Similar to the OPAMP case study, we then take this response, convert it into relation query triplets, and search within AMSnet-KG for matching topologies.

This time, we arrive at two different types of comparator topologies: strong-arm latch comparator and double-tail latch comparator. For tie-breaking purposes, we feed the two names back into the LLM to make a final selection, and arrive at the strongarm latch comparator. The conversations are shown in the Appendix section \ref{appendix:comparator_design} in Fig. \ref{fig:conv_comparator_topology_prompt} through Fig. \ref{fig:conv_comparator_topology_selection}.

\begin{figure*}
    \centering
    \includegraphics[width=0.45\textwidth]{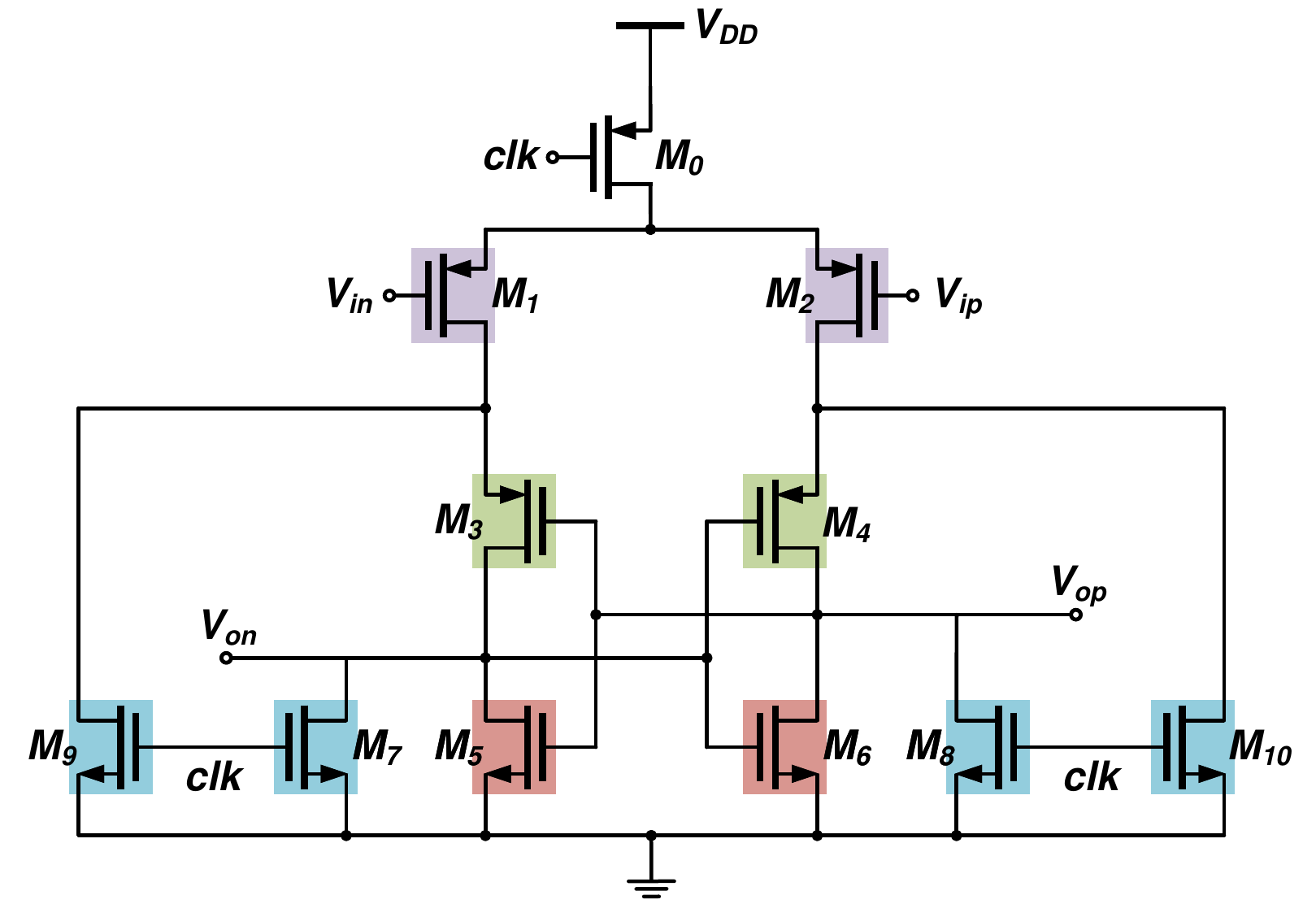}
    \caption{Visualization of the retrieved strong-arm latch comparator, symmetric transistors are highlighted in the same color.}
    \label{fig:full_circuit_comparator}
\end{figure*}

\subsubsection{Sizing}
\begin{equation}
\label{eq:comparator_fom}
\begin{aligned}
FoM =& -\frac{\max(\lg(f_{ov}(x)), -4) - \lg(f_{ov}^{\max})}{\lg(f_{ov}^{\max}) - \lg(f_{ov}^{\min})} - \frac{\max(\lg(f_{pd}(x)), -9) - \lg(f_{pd}^{\max})}{\lg(f_{pd}^{\max}) - \lg(f_{pd}^{\min})} - \frac{\max(\lg(f_{p}(x)), -4) - \lg(f_{p}^{\max})}{\lg(f_{p}^{\max}) - \lg(f_{p}^{\min})}
\end{aligned}
\end{equation}
Similar to the OPAMP case study, we employ Bayesian optimization to obtain the optimal sizing using the FoM in Equation \ref{eq:comparator_fom}, where $f_{ov}(x)$, $f_{pd}(x)$, $f_{p}(x)$ each represent the simulated offset voltage, propagation delay, and power for given sizing $x$. Every metric is viewed on a logarithmic scale, similar to GBW in the OPAMP case study. A similar set of local annotations are used to guide sizing constraints. As shown in Fig. \ref{fig:full_circuit_comparator}, there are a total of 5 PMOS and 6 NMOS transistors. Each of them is parameterized by a length and a width for a total of 22 parameters. After considering symmetry, the number of free parameters drop to 10. 

\begin{figure*}
    \centering
    \includegraphics[width=0.45\textwidth]{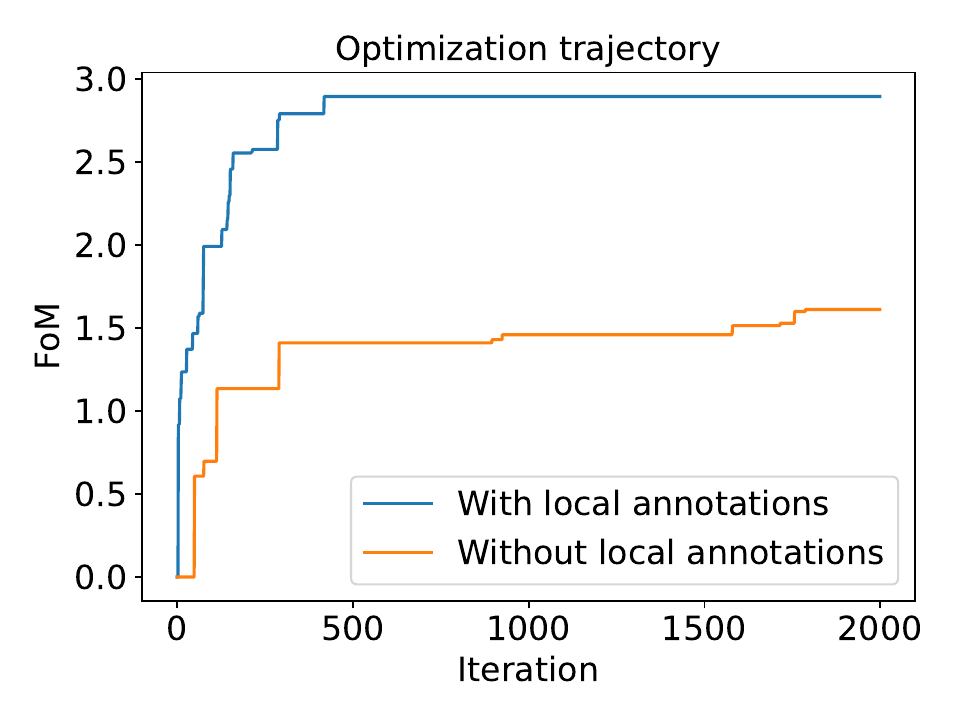}
    \caption{Bayesian optimization trajectory for strong-arm latch comparator}
    \label{fig:bayesian_optimization_fom_comparator_comparator}
\end{figure*}

The optimization trajectory is shown in Fig. \ref{fig:bayesian_optimization_fom_comparator_comparator}. Again similar to the OPAMP case study, having local annotations significantly improves convergence time as well as final performance. As shown in Table \ref{tab:fom_result_cmp}(b), all desired performance specs have been met. This case study provides another data point to prove the generality of AMSgen.

\begin{table}[tb]
\centering
\scriptsize
\resizebox{\columnwidth}{!}{
\setlength{\tabcolsep}{1.2mm}
\begin{tabular}{l|l|l|l|l|l|l}
\toprule
    & Topology & BO constraint & FoM $\uparrow$ & Offset voltage (V)$\downarrow$ & Propagation delay (s)$\downarrow$ & Power (W) \\ \midrule
Specification & - & - & - & 1.00 $\times 10^{-4}$ & 1.00 $\times 10^{-9}$ & 1.00 $\times 10^{-4}$ \\ \midrule
(a) & Strong-arm & W/o local annotations & 1.61 & 7.98 $\times 10^{-4}$ & 5.81 $\times 10^{-10}$ & 3.42 $\times 10^{-4}$ \\ \midrule
(b) & Strong-arm & W/ local annotations & \textbf{2.89} & \textbf{3.50} $\mathbf{\times 10^{-5}}$ & \textbf{1.19} $\mathbf{\times 10^{-11}}$ & \textbf{8.84} $\mathbf{\times 10^{-5}}$ \\ \bottomrule
\end{tabular}
}
\caption{Designs optimized by Bayesian optimization}
\label{tab:fom_result_cmp}
\end{table}

\section{Conclusions and Disscussions}
\label{sec:conclusion}
In this paper, we introduce a new high-quality dataset for AMS circuits, titled AMSnet-KG, which includes schematics, netlists, and manual annotations, all presented as a knowledge graph. We also propose an AMS circuit topology design process based on LLM and KG-RAG. This process begins from input performance specifications into LLM to obtain a design strategy (e.g., circuit architecture), which is then transformed into relation query triplets. Relevant circuit components and testbenches are then retrieved from the knowledge graph. After assembling a complete circuit using retrieved components, parameter sizing is finalized using Bayesian optimization based on design constraints. If the resulting design fails to meet specifications (or costs too much circuit area to meet specifications), the design strategy is adjusted and the topology is refined. We have experimented two case studies and obtained desired OPAMP and Comparator designs.

In the future, we plan to enrich AMSnet-KG with additional information, such as specific parameter sets with corresponding circuit performance. We also plan to introduce new circuits types and additional topologies. The increasing size and dimension could support more application scenarios, such as training AMS-specific foundation models. We will also develop more efficient performance modeling/prediction with reduced SPICE simulation runs or no SPICE simulation for the sizing procedure, and develop better sizing algorithms compared to existing algorithms.

\bibliographystyle{ACM-Reference-Format}
\bibliography{ref}
\clearpage
\appendix
\section{Background}
\label{sec:bo}
\subsection{Bayesian Optimization}
Bayesian optimization (BO) is a strategy for global optimization, particularly suitable for optimizing black-box functions, usually employed when the function evaluation is expensive or the search space is too large to exhaust. BO treats the black-box function to be optimized as a stochastic process, typically modeled using a Gaussian Process (GP) as a surrogate model. The GP provides a flexible way to describe the distribution of the black-box function and updates predictions based on existing data points.

By sampling initial data points, the GP model is trained to fit the black-box function. An acquisition function is selected to balance exploration and exploitation, determining the next sampling point. In the input space, the point that maximizes the acquisition function is found. The selected new point is then evaluated through simulation, and its result is added to the existing dataset, updating the GP model. By balancing exploration and exploitation, BO can avoid getting trapped in local optima and find the global optimum.

\subsection{Gaussian Process}
A GP is a stochastic process defined over an input space, where any finite subset of random variables follows a multivariate Gaussian distribution. Given a training dataset \( \{(x_i, y_i)\}_{i=1}^n \), where \( x_i \) are the inputs and \( y_i \) are the target values. A GP is completely defined by its mean function \( \mu(x) \) and covariance function \( k(x, x') \). The mean function represents the expected value of the function at a given input, and the covariance function (also called the kernel function) represents the correlation or similarity between any two points. Generally, the mean function of a GP is assumed to be zero:
\begin{equation}
\begin{aligned}
m(x) = \mathbb{E}[f(x)] = 0
\end{aligned}
\end{equation}
The kernel function \( k(x, x') \) is defined as:
\begin{equation}
\begin{aligned}
k(x, x') = \mathbb{E}[(f(x) - m(x))(f(x') - m(x'))]
\end{aligned}
\end{equation}
For a new input point \( x_* \), the predictive distribution is also a Gaussian distribution. The mean and variance of the prediction can be computed as follows:
\begin{equation}
\begin{aligned}
\mu_* = k(x_*, X)^T K^{-1} y
\end{aligned}
\end{equation}
\begin{equation}
\begin{aligned}
\sigma^2_* = k(x_*, x_*) - k(x_*, X)^T K^{-1} k(x_*, X)
\end{aligned}
\end{equation}
where \( k(x_*, X) \) is the covariance vector between the new input point and all training data points, and \( y \) is the vector of target values for the training data.

\subsection{Acquisition Function}
The acquisition function is a key component in BO, used to select the next evaluation point. It balances exploration and exploitation, making trade-offs between exploring unknown regions and exploiting the known best regions. 

\textbf{Expected Improvement (EI)}:
EI measures the expected amount of improvement over the current best point. The formula is:
\begin{equation}
\alpha_{EI}(x) = \mathbb{E}[\max(0, f(x) - f(x^+))]
\end{equation}
where \( f(x^+) \) is the current best observed value.

By modeling our custom FoM as a black-box function of the design parameters, we are able to use a GP to fit the simulation and use BO to optimize our parameters without explicit knowledge.

\section{Complete LLM Conversations}
\label{appendix:llm_conversations}
\subsection{Opamp Initial Design}
\label{appendix:opamp_initial_design}
\begin{figure}[htb]
    \centering
    \includegraphics[width=\textwidth]{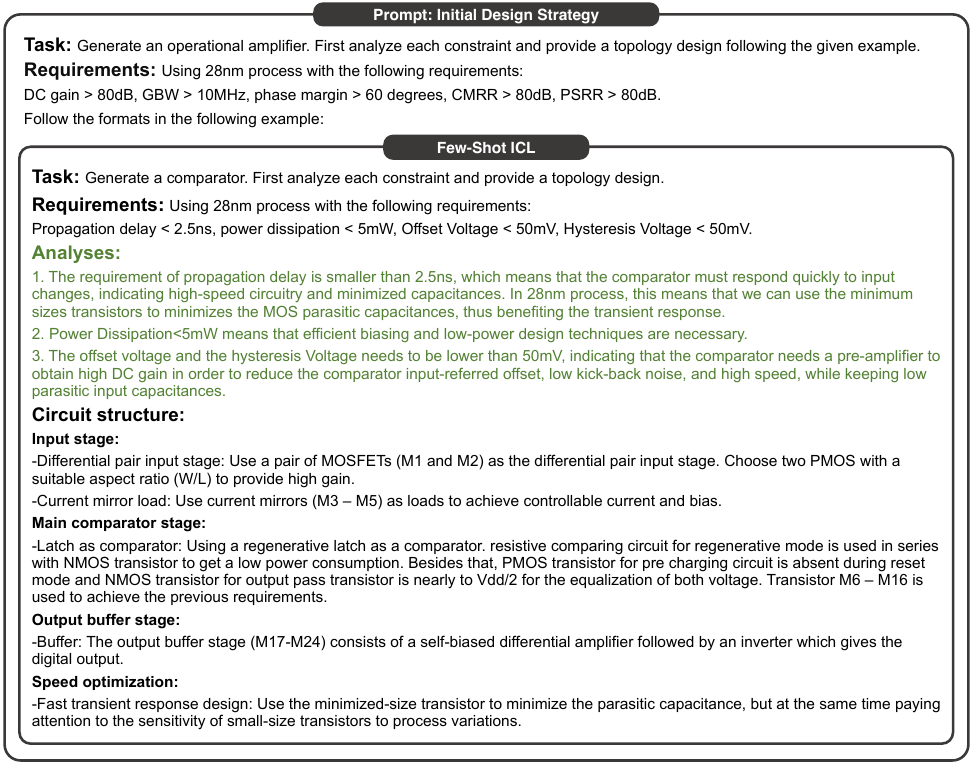}
    \caption{LLM prompt: from performance specification to initial design strategy}
    \label{fig:prompt_01_example}
\end{figure}

\begin{figure}[htb]
    \centering
    \includegraphics[width=\textwidth]{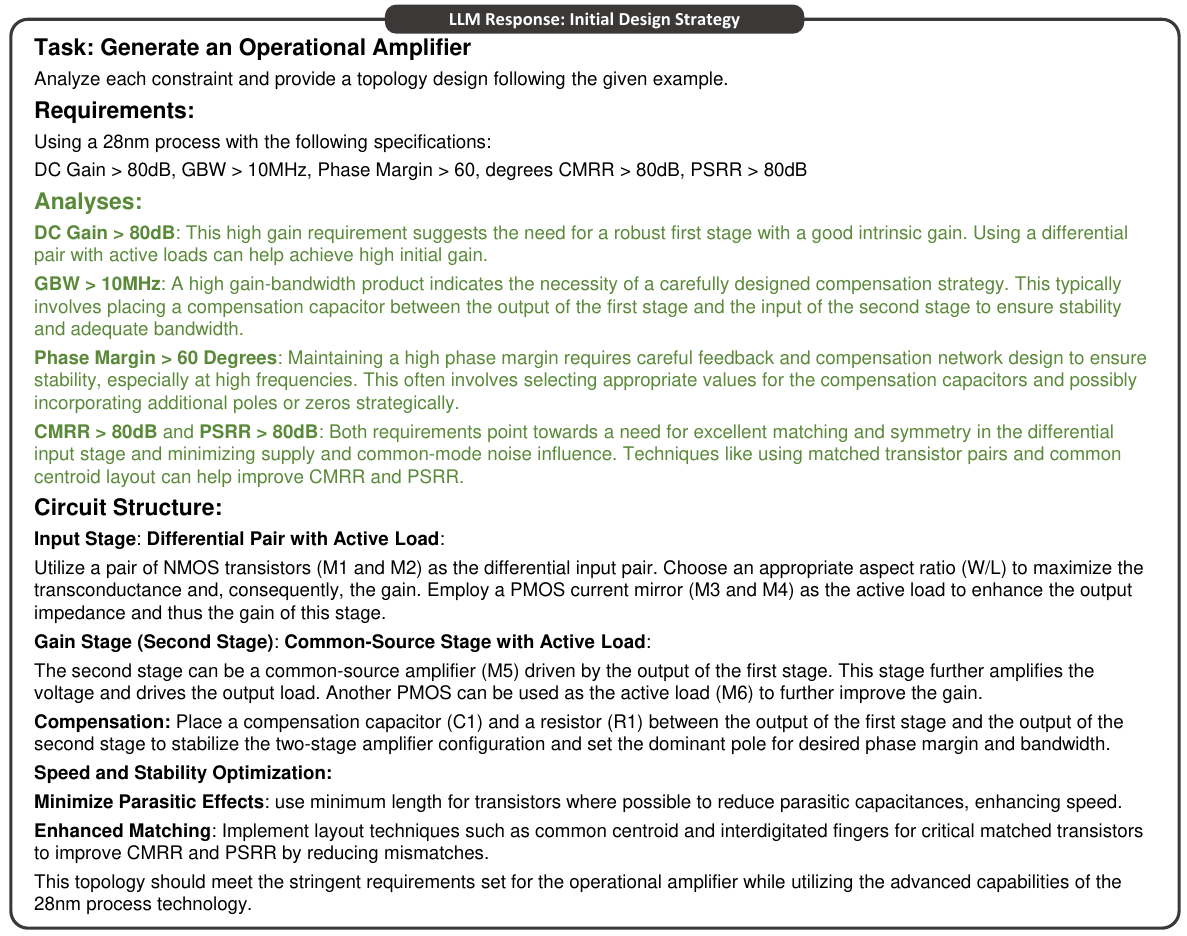}
    \caption{LLM response: from performance specification to initial design strategy}
    \label{fig:prompt_01_response}
\end{figure}

\begin{figure}[htb]
    \centering
    \includegraphics[width=\textwidth]{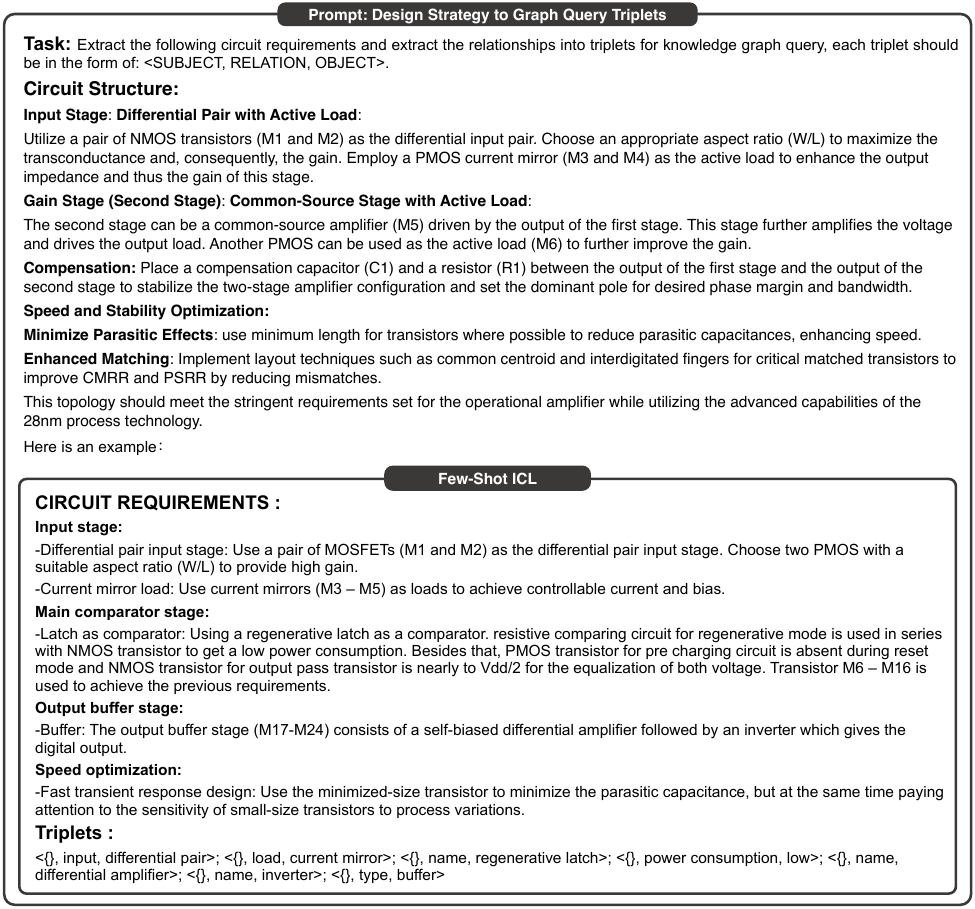}
    \caption{LLM prompt: from initial design strategy to graph query triplets}
    \label{fig:prompt_02_example}
\end{figure}

\begin{figure}[htb]
    \centering
    \includegraphics[width=\textwidth]{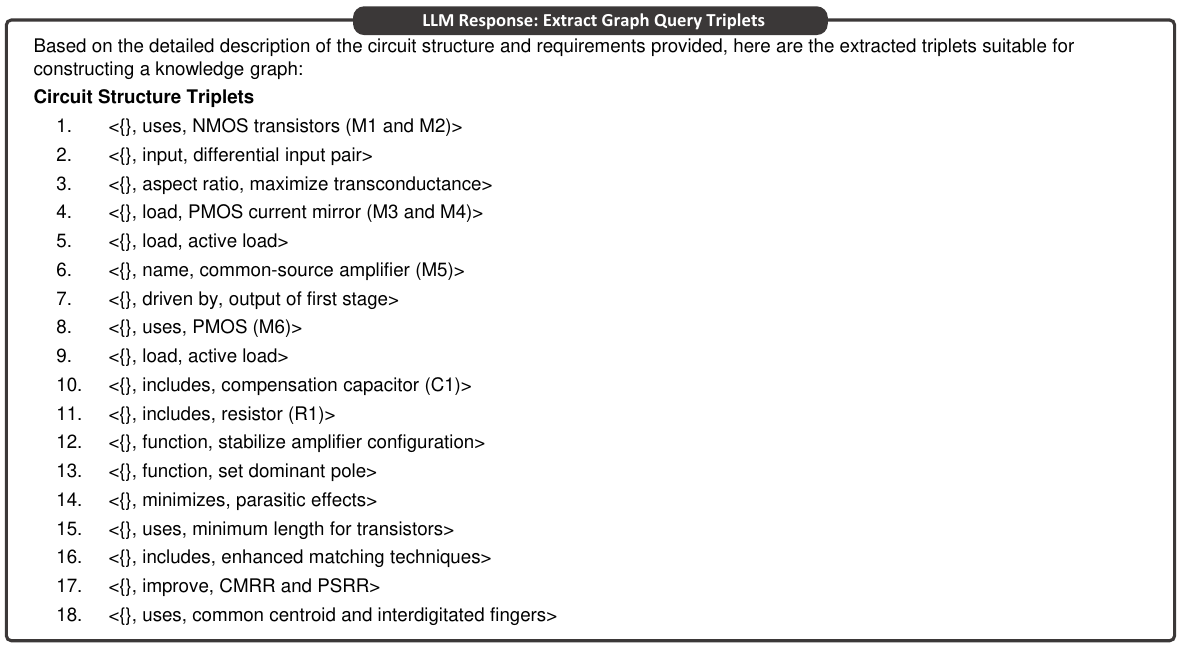}
    \caption{LLM response: from initial design strategy to graph query triplets}
    \label{fig:prompt_02_response}
\end{figure}

\clearpage
\subsection{Opamp Regenerated Design}
\label{appendix:opamp_regenerated_design}
\begin{figure}[htb]
    \centering
    \includegraphics[width=\textwidth]{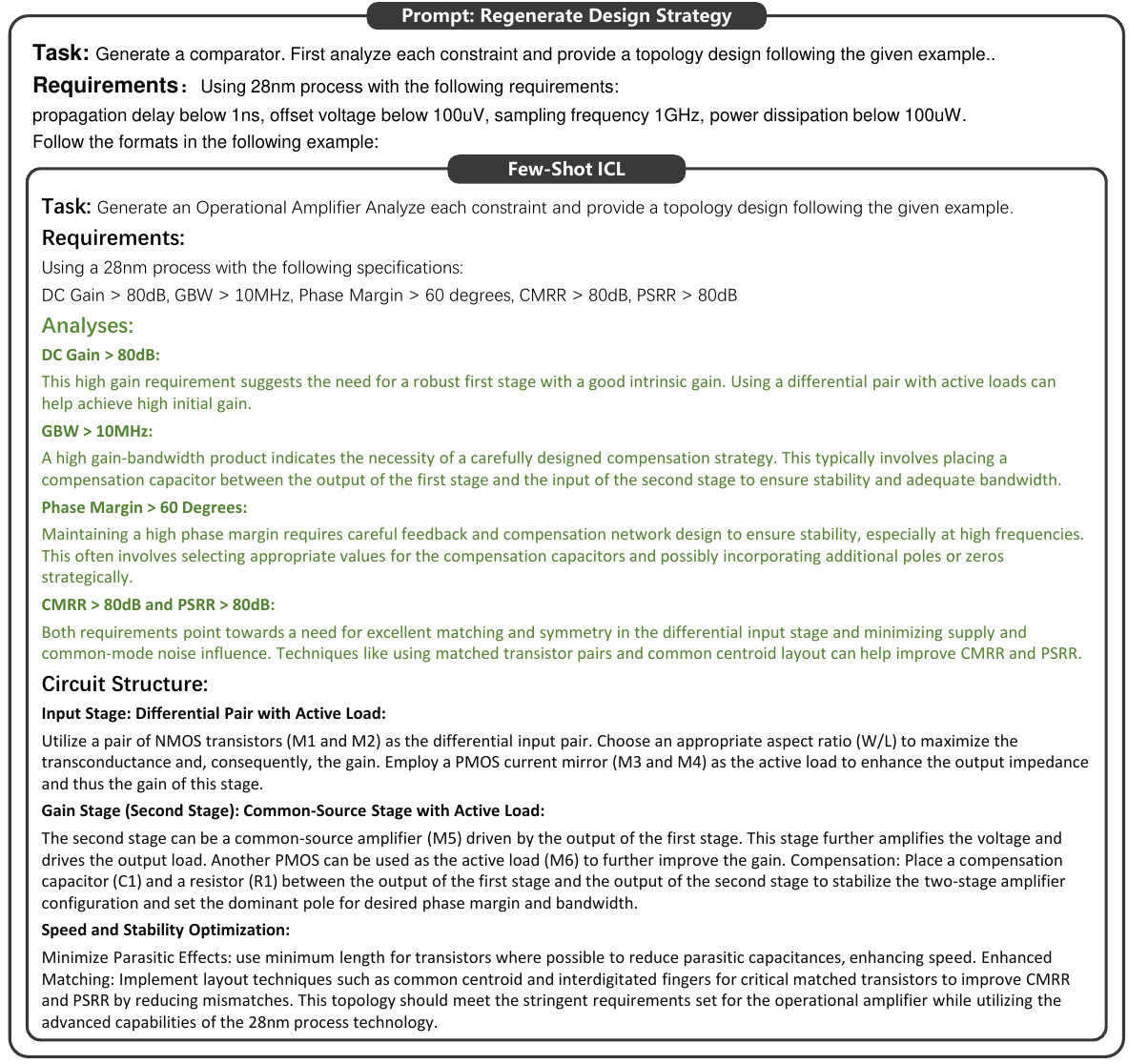}
    \caption{LLM prompt: from performance specification to revised design strategy}
    \label{fig:prompt_03_example}
\end{figure}

\begin{figure}[htb]
    \centering
    \includegraphics[width=\textwidth]{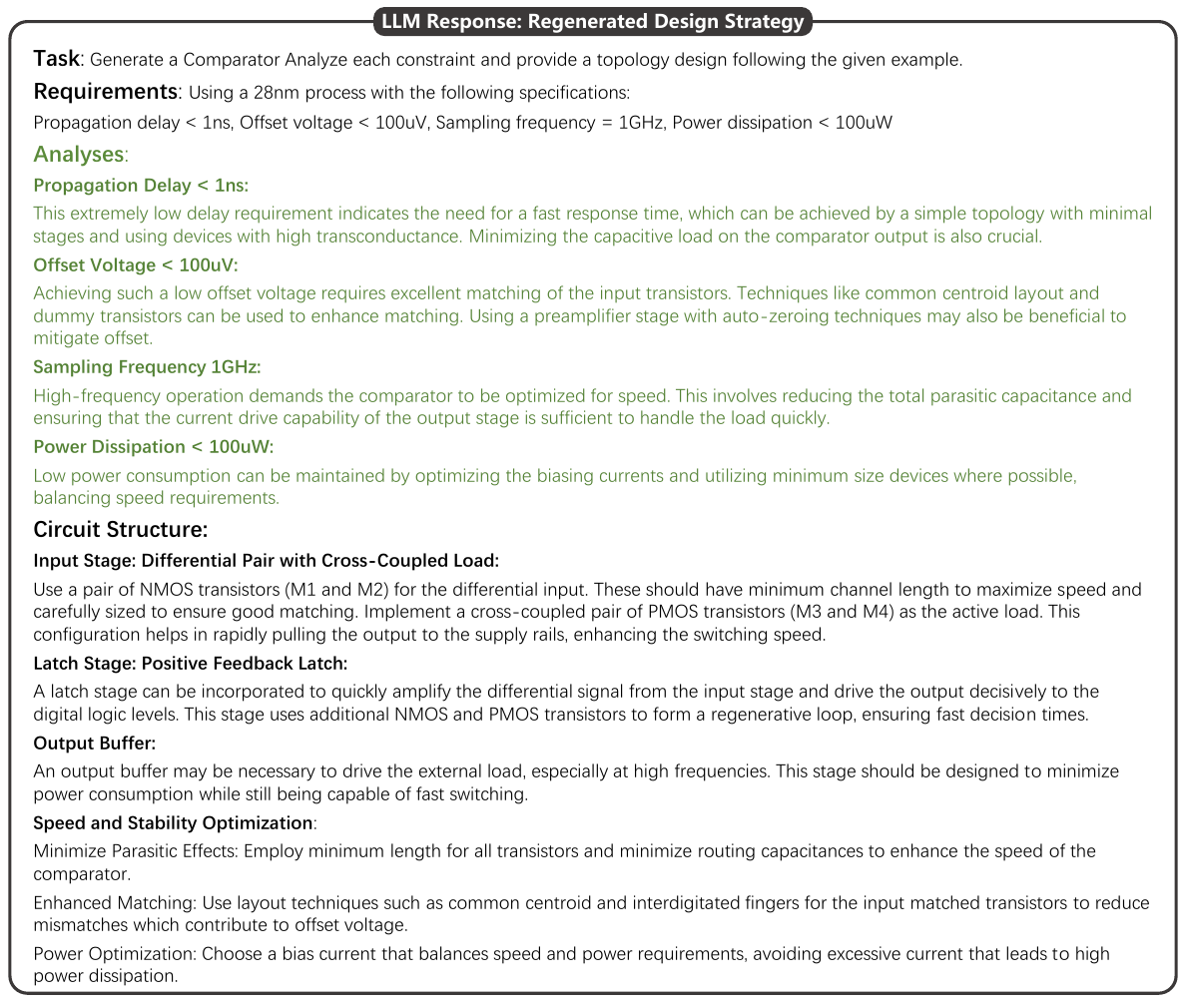}
    \caption{LLM response: from performance specification to revised design strategy}
    \label{fig:prompt_03_response}
\end{figure}

\begin{figure}[htb]
    \centering
    \includegraphics[width=\textwidth]{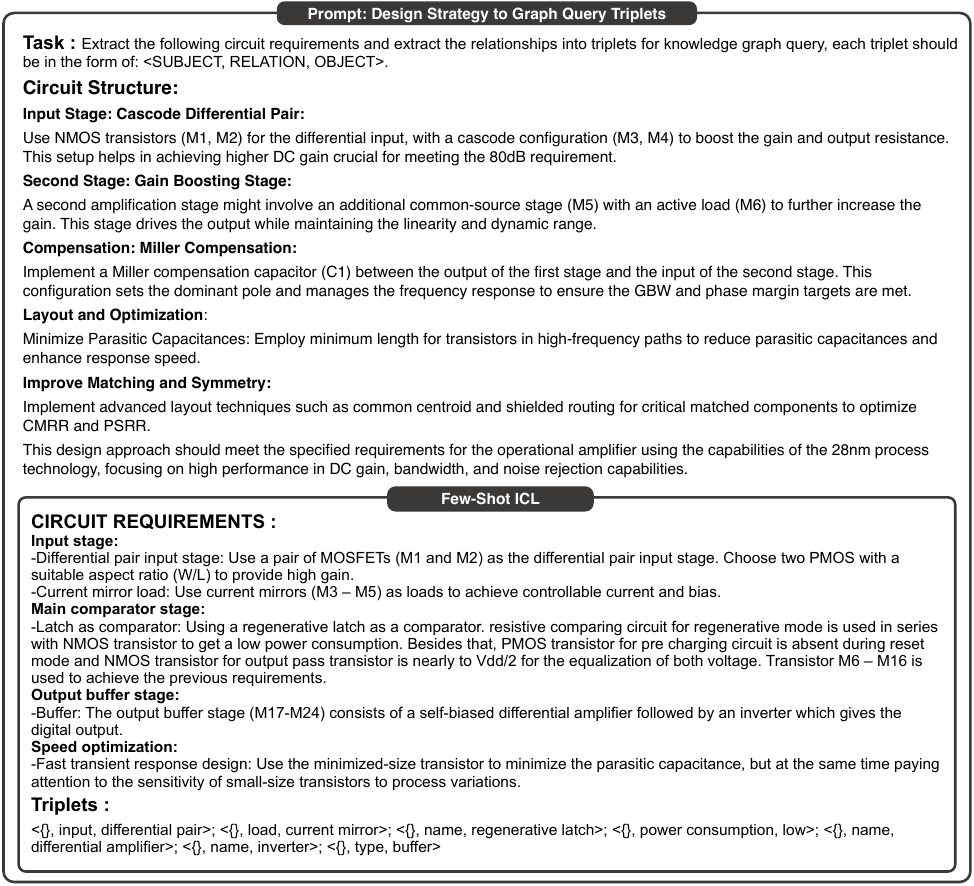}
    \caption{LLM prompt: from revised design strategy to graph query triplets}
    \label{fig:prompt_04_example}
\end{figure}

\begin{figure}[htb]
    \centering
    \includegraphics[width=\textwidth]{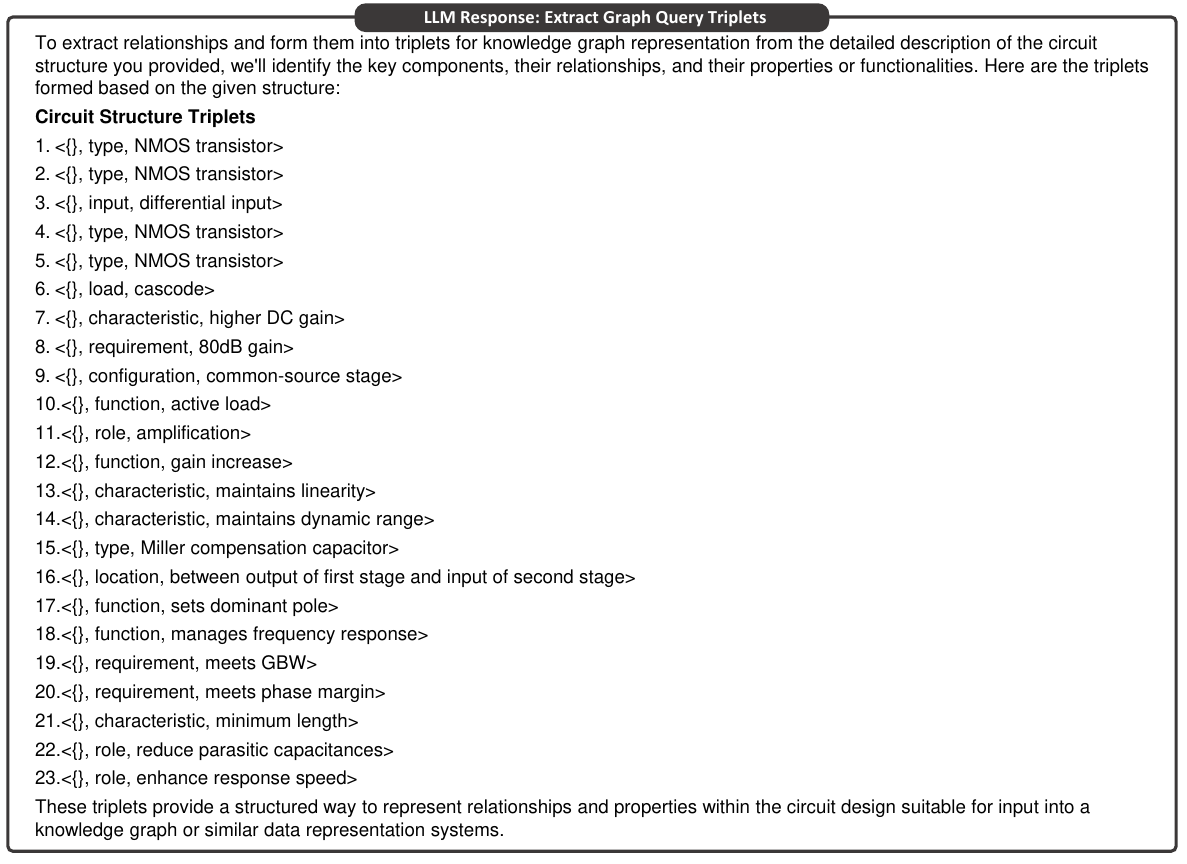}
    \caption{LLM response: from revised design strategy to graph query triplets}
    \label{fig:prompt_04_response}
\end{figure}

\clearpage
\subsection{Comparator Design}
\label{appendix:comparator_design}
\begin{figure}[htb]
    \centering
    \includegraphics[width=\textwidth]{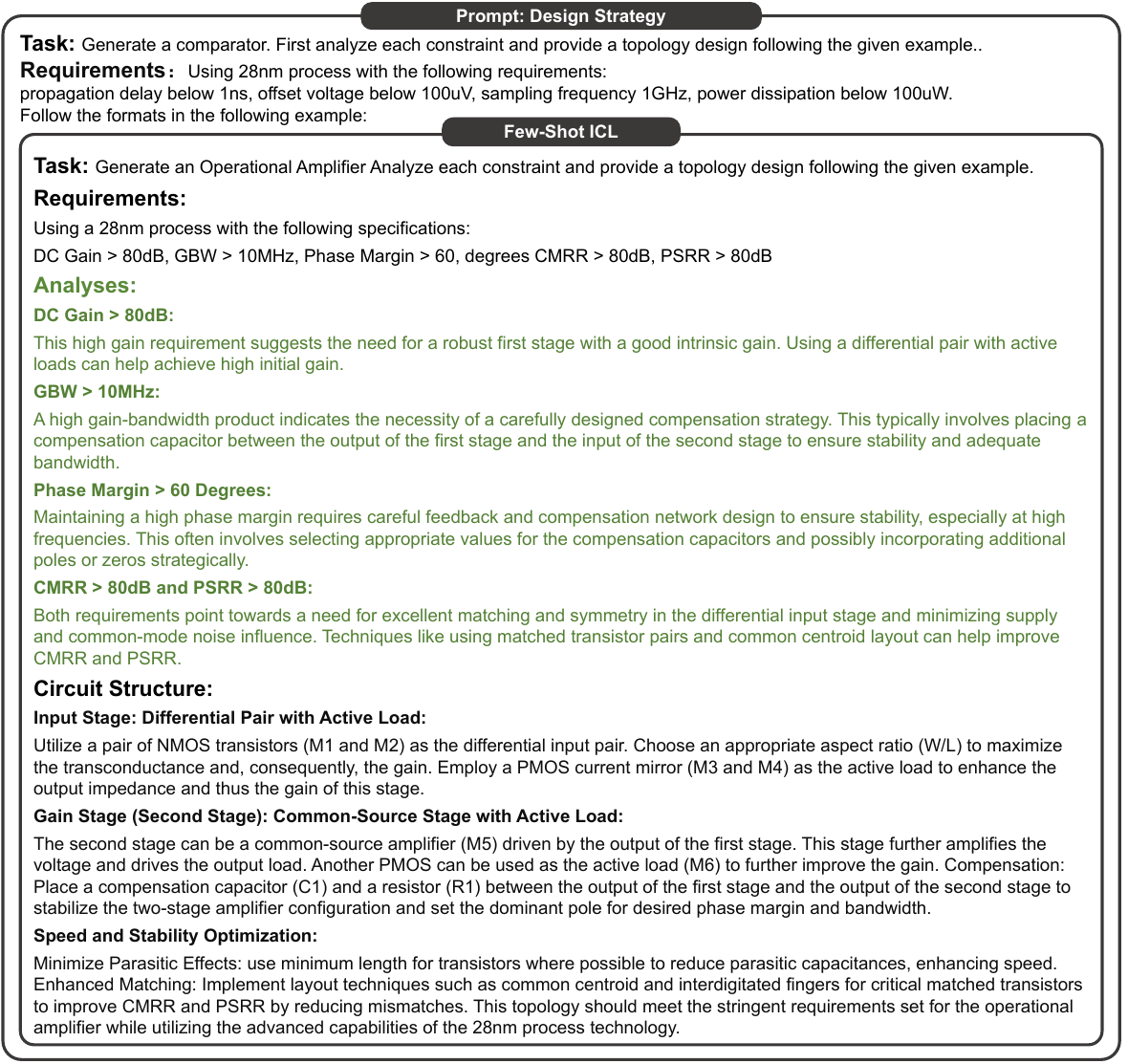}
    \caption{LLM prompt: from performance specification to design strategy}
    \label{fig:conv_comparator_topology_prompt}
\end{figure}

\begin{figure}[htb]
    \centering
    \includegraphics[width=\textwidth]{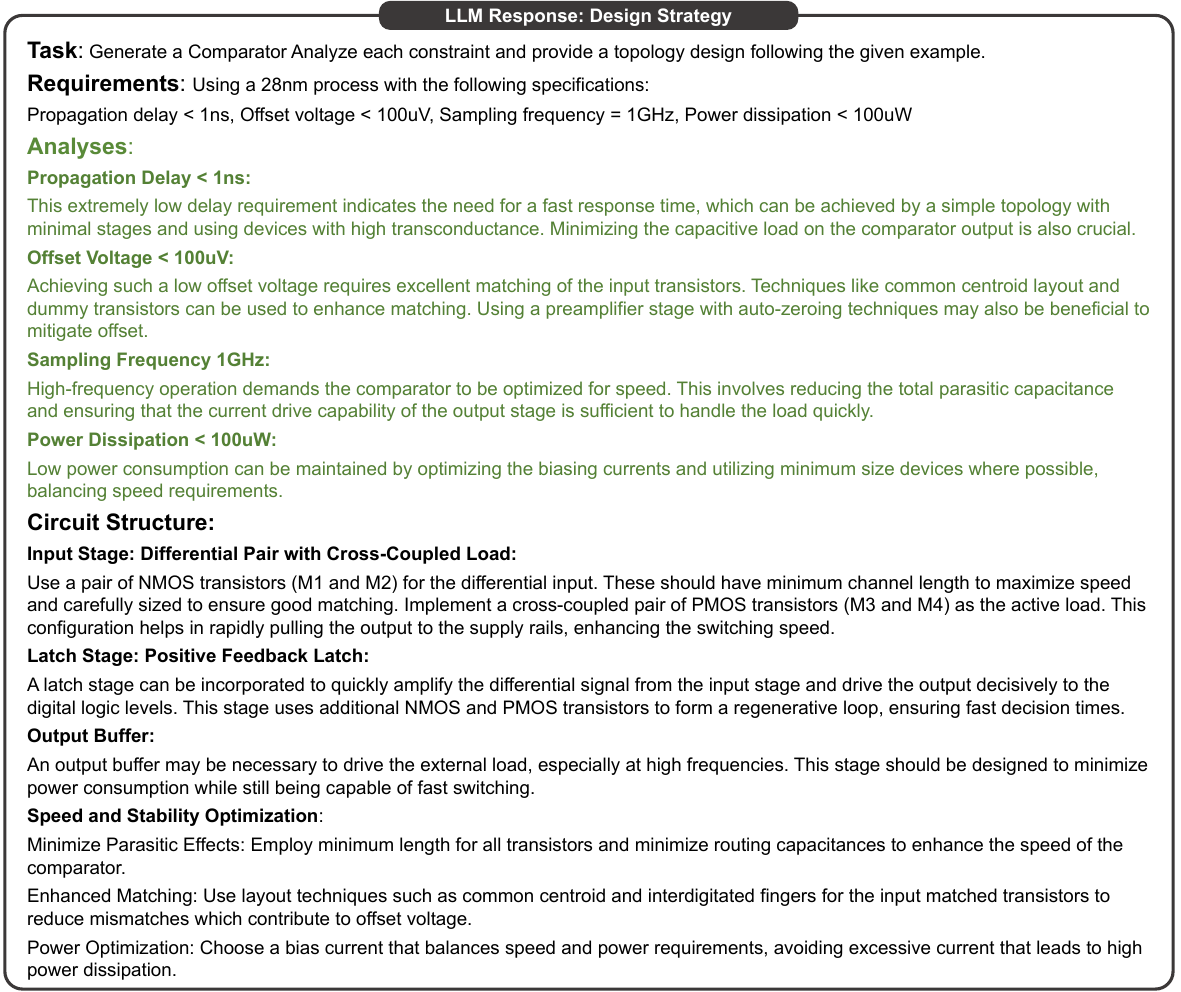}
    \caption{LLM response: from performance specification to design strategy}
    \label{fig:conv_comparator_topology_response}
\end{figure}

\begin{figure}[htb]
    \centering
    \includegraphics[width=\textwidth]{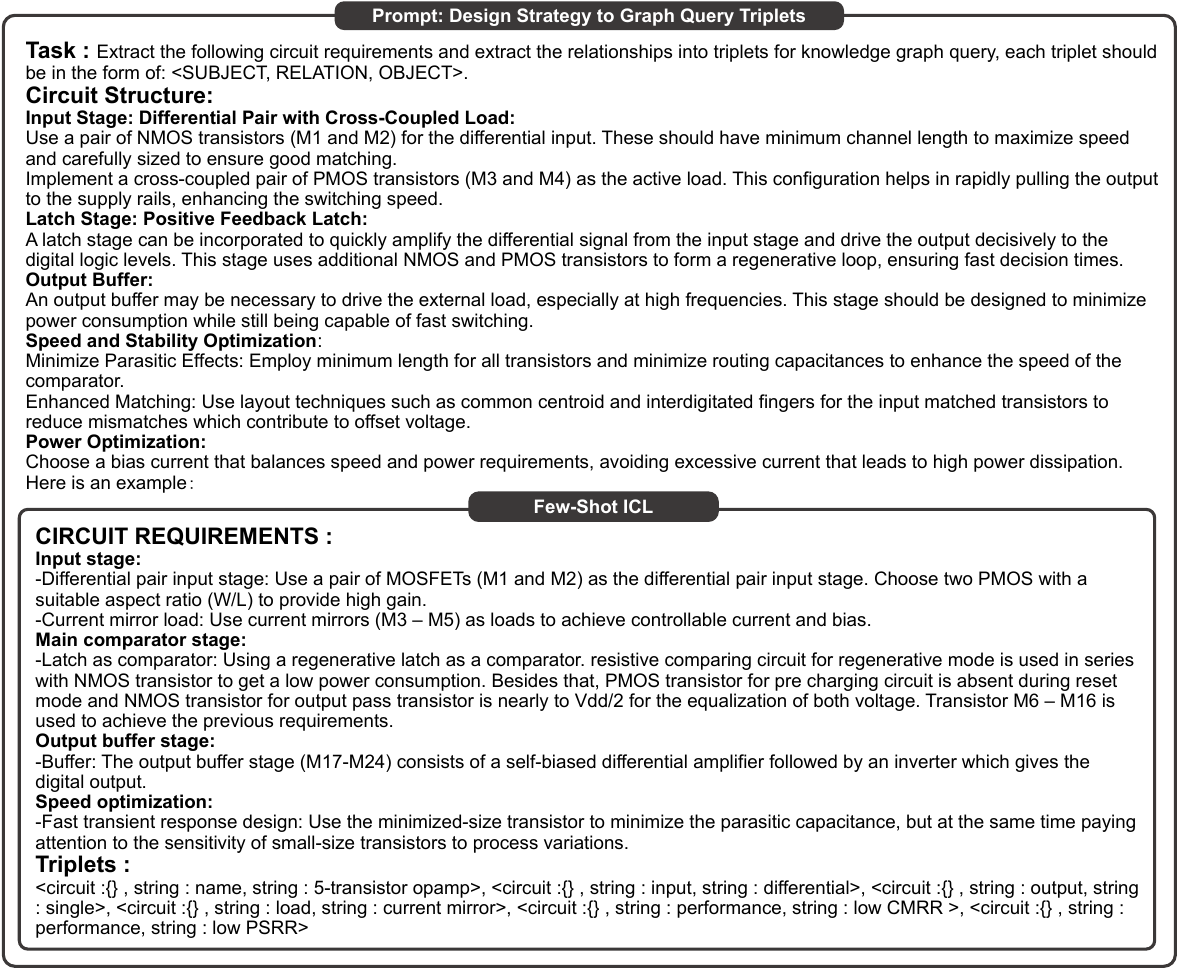}
    \caption{LLM prompt: from design strategy to graph query triplets}
    \label{fig:conv_comparator_kg_prompt}
\end{figure}

\begin{figure}[htb]
    \centering
    \includegraphics[width=\textwidth]{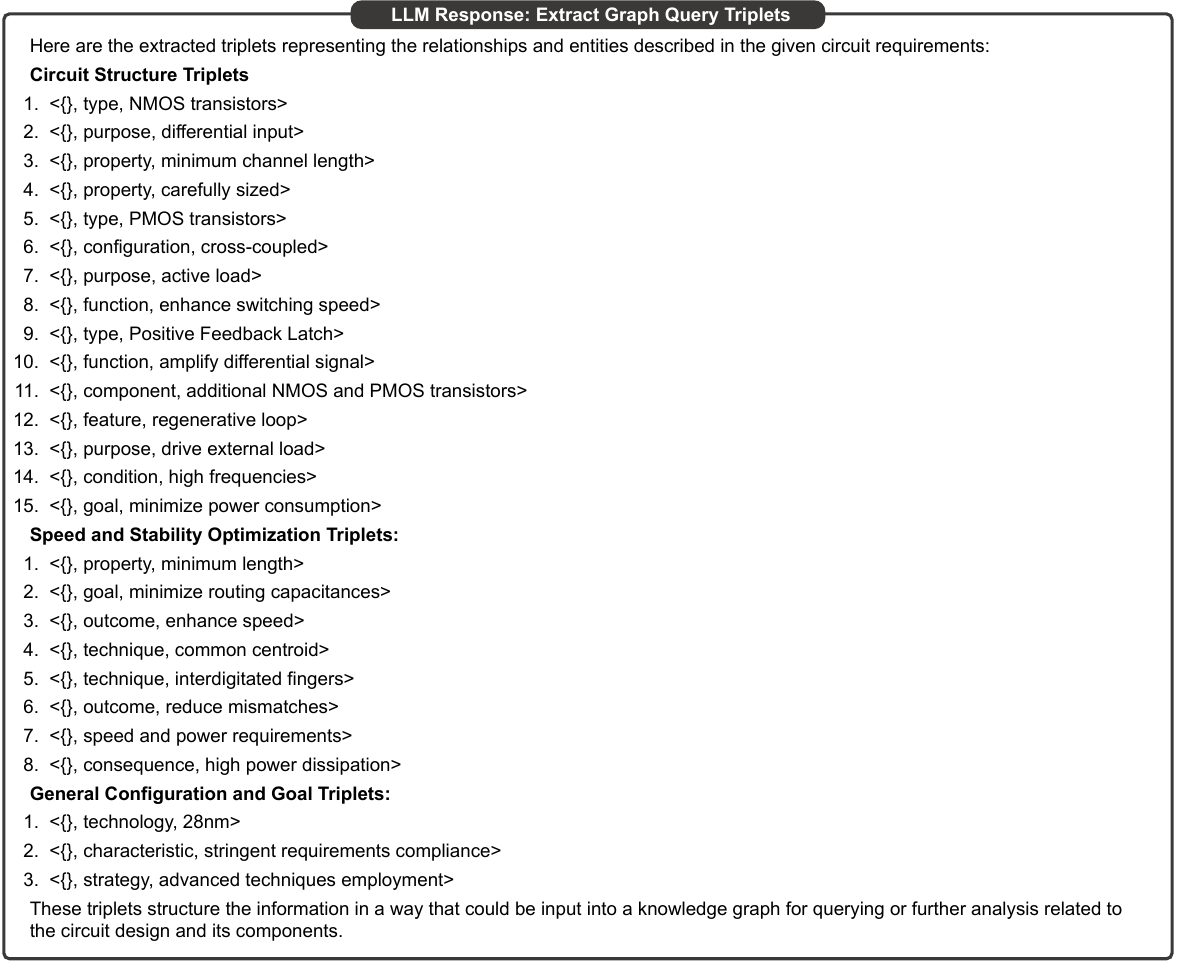}
    \caption{LLM response: from design strategy to graph query triplets}
    \label{fig:conv_comparator_kg_response}
\end{figure}

\begin{figure}[htb]
    \centering
    \includegraphics[width=\textwidth]{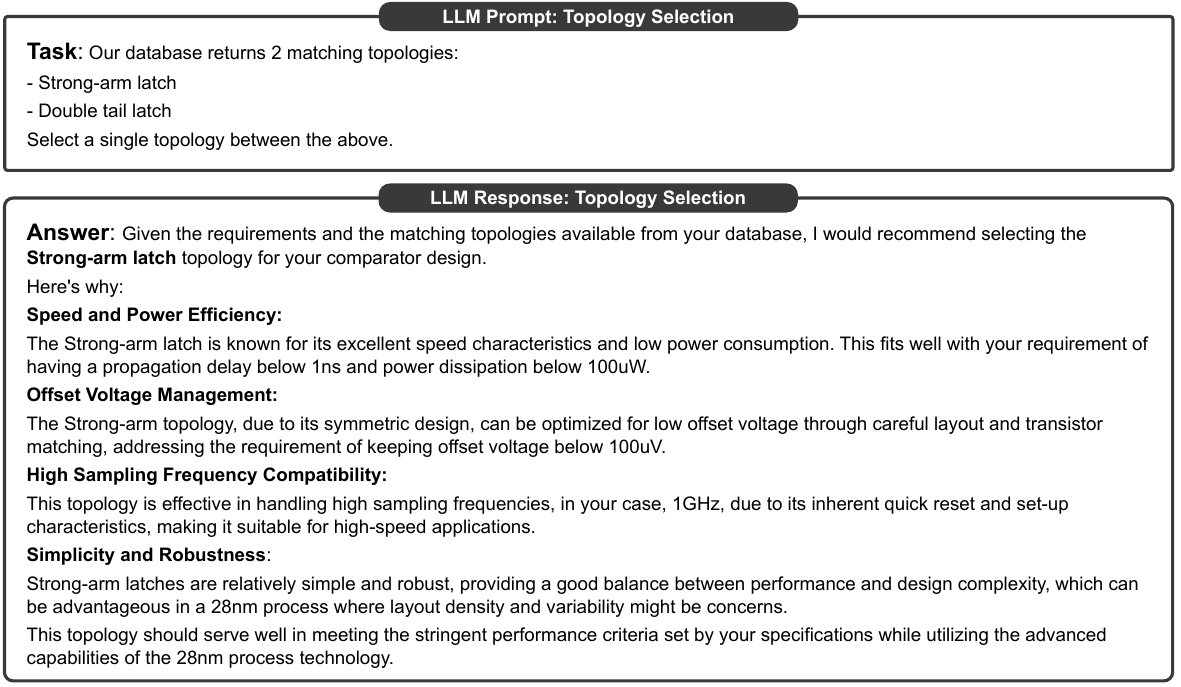}
    \caption{LLM prompt and response: tie breaker between valid topologies, continuing on the context of Fig. \ref{fig:conv_comparator_topology_response}}
    \label{fig:conv_comparator_topology_selection}
\end{figure}

\end{document}